%% file: 0-main.tex
\documentclass[journal=gmj]{CUP-JNL-DTM}%

\usepackage{graphicx}
\usepackage{multicol,multirow}
\usepackage{amsmath,amssymb,amsfonts}
\usepackage{mathrsfs}
\usepackage{amsthm}
\usepackage{rotating}
\usepackage{appendix}
\usepackage{ifpdf}
\usepackage[T1]{fontenc}
\usepackage{newtxtext}
\usepackage{newtxmath}
\usepackage{textcomp}
\usepackage{amsthm}
\usepackage{xcolor}
\usepackage{tabularx}
\usepackage{array}
\usepackage{lipsum}
\usepackage[colorlinks,allcolors=blue]{hyperref}

\theoremstyle{definition}

\numberwithin{equation}{section}

\jname{Research Article}
\jyear{2026}
\begin{document}

\pagenumbering{gobble}
\pagestyle{empty}

\begin{Frontmatter}

\title[Concept-Grounded Attention]{Concept-Grounded Attention: A Controlled Evaluation of Graph-Injected Attention, Temporal Versioning, and Epistemic Status}

% There is no need to include ORCID IDs in your .pdf; this information is captured by the submission portal when a manuscript is submitted. 
\author[1]{Sachin Dev Duggal}
\author[1]{Pradyumna Swarnalatha Ramanna}\orcid{0009-0002-3852-9068}
\author[1]{Alexandros Vassiliades}\orcid{0000-0003-4569-503X}

\authormark{Sachin Dev Duggal \textit{et al}.}

\address[1]{\orgname{SeKondBrain AI Labs}, \orgaddress{\city{London}, \country{United Kingdom}}}

\authormark{Sachin Dev Duggal et al.}


\abstract{Knowledge-intensive language-model systems typically represent external knowledge as text chunks or static graphs, with limited support for concept evolution, point-in-time reasoning, and distinctions between validated and inferred knowledge. We introduce the Concept Lifecycle Model (CLM), which represents concepts as persistent, graph-grounded, temporally versioned entities with explicit provenance and epistemic status, and Concept-Grounded Attention (CGA), which injects concept-graph structure into transformer computation through graph-biased self-attention (Form A) and gated cross-attention over concept nodes (Form B). We evaluate the framework in three controlled settings and pair every internal diagnostic with the same measurement under a disabled mechanism. On 200 MuSiQue and HotpotQA questions with retrieval held fixed, concept-graph retrieval recovers explicit multi-hop paths but does not improve evidence recall. Form A appears to steer attention, allocating 2.76 times more attention to gold than to distractor concepts, but the same ratio is obtained with Form A disabled: it is a property of the pretrained model, the learned bias is about four orders of magnitude smaller than the attention logits, and no answer changes. The collapse of Form B (F1 0.006) is traced to a normalisation that does not preserve the pretrained computation. An identity-preserving Form B improves F1 from 0.188 to 0.221, but concepts taken from other questions yield 0.213, so most of the gain reflects added capacity; a borderline concept-specific gain with a cross-encoder relevance prior (0.010 F1, $p=0.05$) does not replicate with flan-t5-base, and a query-conditioned Form A gains no more than a shuffled control. On LongMemEval, representing time improves answer accuracy by 13 to 25 points for every generator tested, up to 122 billion parameters, while simplified CLM version resolution performs on a par with dated serialisation because concept identity is not established reliably. On a synthetic graph requiring judgements of source independence, protocol-derived epistemic status reduces unsupported assertions from 28\% to 0.1\% in a fine-tuned small model and from 19--68\% to 0--5\% in 72--122-billion-parameter models that are told the confirmation rule but must apply it themselves. The results support the CLM's premise that temporal validity and epistemic status should be explicit more than the proposed mechanisms for injecting graph structure into attention, and they show that attention diagnostics for knowledge-enhanced models are uninformative without a disabled-mechanism control.
}

\end{Frontmatter}

\input{1-introduction}
\input{2-related}

\input{3-methodology}
\input{4-evaluation}

\input{5-results}
\input{6-discussion}
\input{7-conclusion}

\section*{Acknowledgment}
The authors gratefully acknowledge Benjamin Brey (SeKondBrain), Priyanka Kochhar (SeKondbrain) \& Dr Sharon Jheeta (outside advisor) for reading early versions of this paper and providing insightful editorial suggestions.

\appendix
\input{8-appendix}

\begin{Backmatter}

% \section{Results}
% \subsection{Multi-Hop Reasoning}
% \subsection{Temporal Reasoning}
% \subsection{Epistemic Calibration}
% \subsection{Ablation Results}
% \subsection{Efficiency Analysis}

\bibliography{bibo}

\end{Backmatter}

\end{document}

%% file: 1-introduction.tex
\section{Introduction}\label{sec:intro}

Large Language Models (LLMs) have substantially advanced natural language understanding and generation, yet their ability to reason reliably over evolving external knowledge remains limited. Retrieval-Augmented Generation (RAG) improves factual grounding by supplying external evidence at inference time, while more recent graph-based approaches introduce entities, relations, and multi-hop retrieval into the generation process \cite{lewis2020retrieval,edge2024graphrag,gutierrez2024hipporag}. However, most existing systems continue to operate primarily over tokens, text chunks, documents, or static graph representations. Consequently, they provide limited support for queries that require historically accurate knowledge, explicit provenance, concept evolution, or distinctions between validated and merely inferred claims.

These limitations are particularly important in knowledge-intensive domains such as law, finance, healthcare, regulation, and scientific research. In such settings, a concept may be distributed across multiple documents, change meaning over time, become superseded by a new interpretation, or remain uncertain until independently confirmed. Increasing the context-window size does not directly address these requirements, because it expands the amount of available text without providing persistent concept identity, temporal validity, structured relations, or explicit epistemic status. Similarly, graph-based retrieval can identify relevant entities and relationships, but the retrieved structure is often serialized into text before language-model inference. This transformation limits the extent to which graph topology, temporal validity, and evidence quality can directly influence the model's internal computation.

This paper investigates an alternative premise: concepts, rather than tokens or documents, should serve as the primary units of knowledge-intensive reasoning. We define a concept as a persistent, graph-grounded, and temporally versioned entity whose meaning is determined jointly by its semantic representation, its supporting evidence, and its relational position within a dynamic knowledge graph. Documents are therefore treated as evidence supporting concepts rather than as the primary objects over which reasoning is performed.

Building on this view, we introduce the Concept Lifecycle Model (CLM), a formal model describing how concepts emerge, evolve, split, become superseded, are invalidated, and are archived while preserving their temporal history and provenance. The model assigns a stable identity to each concept and maintains versioned representations that support point-in-time resolution. It also introduces a four-tier epistemic model distinguishing validated, hypothesised, promoted, and invalidated relationships. Epistemic status is determined by the confirmation protocol and the independence of the supporting evidence rather than by the confidence score of a single statistical model.

We further propose Concept-Grounded Attention (CGA), a neuro-symbolic transformer architecture that integrates concept-graph information directly into model computation. CGA combines two complementary mechanisms. First, graph-biased self-attention introduces structural, temporal, and epistemic information as biases within token-level attention. Second, cross-attention over retrieved concept nodes gives the language model direct access to a temporally scoped conceptual subgraph. This dual-attention design enables graph structure to influence generation without first being reduced to flat textual context.

The broader CGA pipeline combines vector similarity retrieval, multi-hop graph traversal, and temporally scoped subgraph retrieval. An adjudication mechanism selects or fuses the resulting evidence according to the query requirements. Queries with explicit temporal anchors can therefore be resolved against historical concept states, while queries requiring relational reasoning can use multi-hop graph structure. Epistemic weights further control the influence of validated, hypothesised, promoted, and invalidated relationships during reasoning.

A central question is whether such structural signals change what the model generates, or only how it distributes attention internally. We evaluate the structural components of CGA on multi-hop question answering under a controlled design in which all compared systems receive exactly the same retrieved evidence, and we pair every internal diagnostic with the same measurement taken with the mechanism disabled. We further evaluate CLM point-in-time resolution on a public temporal benchmark and protocol-derived epistemic status on a controlled synthetic graph.

The main contributions of this work are as follows:

\begin{itemize}
    \item \textbf{Concept Lifecycle Model.} We introduce a formal model of concept emergence, gradual evolution, discontinuous change, hierarchical splitting, archiving, and invalidation, together with a four-tier epistemic representation and point-in-time resolution.

    \item \textbf{Concept-Grounded Attention.} We propose a neuro-symbolic architecture that integrates concept-graph structure into transformer computation through graph-biased self-attention (Form A) and gated cross-attention over concept nodes (Form B), and we specify the condition, identity preservation at initialisation, under which such modules can be inserted into pretrained transformers.

    \item \textbf{A controlled negative result for graph-biased attention.} We show that the gold-to-distractor attention preference observed with Form A enabled is reproduced exactly with Form A disabled, quantify the gap between the learned bias and the attention logits, and show that stronger bias disrupts rather than improves generation.

    \item \textbf{Diagnosis and repair of concept cross-attention.} We trace the collapse of Form B to a non-identity-preserving normalisation and show, with mismatched-concept controls, that most of the answer-level gain of a repaired Form B comes from added trainable capacity rather than from concept information.

    \item \textbf{Evaluation of temporal and epistemic representation.} We show that representing time improves temporal question answering on LongMemEval for generators from 780 million to 122 billion parameters, that simplified version resolution does not exceed dated serialisation without reliable concept identity, and that protocol-derived epistemic status nearly eliminates unsupported assertions where source independence must be judged, including for 72--122-billion-parameter models that are given the rule explicitly.
\end{itemize}

The adjudicator, the full lifecycle dynamics of emergence, drift, and splitting, large-scale graph sharding, automated verification of evidence-source independence in real corpora, and distributed synchronisation in multi-agent settings remain outside the scope of the evaluation.

The remainder of this paper is organized as follows. Section~\ref{sec:related} reviews related work on retrieval-augmented generation, graph-based language-model reasoning, graph transformers, temporal knowledge graphs, and neuro-symbolic AI. Section~\ref{sec:methodology} presents the proposed methodology, including the Concept Lifecycle Model, concept-graph construction, retrieval and adjudication, and the Concept-Grounded Attention architecture. Section~\ref{sec:evaluation} describes the research questions, datasets, baselines, evaluation metrics, implementation details, and ablation studies. Section~\ref{sec:results} presents the experimental results and associated statistical analysis. Section~\ref{sec:discussion} interprets the findings and discusses limitations and future work. Finally, Section~\ref{sec:conclusion} concludes the paper and outlines directions for future work.

%% file: 2-related.tex
\section{Related Work}
\label{sec:related}

The proposed framework lies at the intersection of retrieval-augmented generation, graph-based language-model reasoning, graph transformers, temporal knowledge graphs, and uncertainty-aware knowledge representation. Existing work has progressively moved from retrieving flat textual passages toward retrieving and reasoning over structured knowledge. However, these research directions typically address retrieval, structural reasoning, temporal modelling, or uncertainty separately. This section reviews the most relevant developments and positions the CLM and CGA in relation to them.

\subsection{Retrieval-Augmented Generation}

RAG combines parametric language models with non-parametric external knowledge by retrieving relevant passages and conditioning generation on the retrieved evidence \cite{lewis2020retrieval}. This architecture improves access to updated or domain-specific information without requiring the underlying language model to be retrained. Subsequent work has extended the basic RAG pipeline through improved indexing, retrieval, generation, and evaluation strategies \cite{gao2024retrieval}. Nevertheless, conventional RAG systems generally retrieve semantically similar passages independently and represent the retrieved evidence as flat textual context. Relationships distributed across multiple documents therefore remain implicit and must be reconstructed by the language model during generation.

Several approaches improve the adaptability or organization of the retrieval process. Self-RAG trains a language model to determine when retrieval is necessary and to critique the relevance and factual support of its own generations through reflection tokens \cite{asai2024selfrag}. Corrective and adaptive retrieval methods similarly attempt to detect insufficient evidence and invoke additional retrieval when required. These approaches improve retrieval control and answer grounding, but they do not maintain persistent concept identities, temporal histories, or explicit relationships among retrieved knowledge units.

RAPTOR organizes text into a recursively constructed hierarchy of clusters and abstractive summaries, allowing retrieval at multiple levels of granularity \cite{sarthi2024raptor}. This improves reasoning over information distributed across long document collections, but the resulting hierarchy remains primarily a document-abstraction structure rather than a persistent model of concept identity and evolution. HippoRAG instead constructs an entity-based graph and applies Personalized PageRank to support associative retrieval and multi-hop reasoning \cite{gutierrez2024hipporag}. HippoRAG~2 extends this direction toward non-parametric continual learning by combining graph-based associative retrieval with stronger passage integration and online language-model recognition \cite{gutierrez2025ragmemory}. These systems demonstrate the value of structured long-term memory, although their graph structures primarily organize retrieval and do not explicitly represent concept lifecycles, point-in-time states, or protocol-derived epistemic statuses.

Structure-aware retrieval makes a related observation from a different direction. STAIR supplies the complete table of contents of a book in the prompt of a fine-tuned language model and trains it to generate the title of the leaf section that answers a query, using constrained generation to restrict outputs to valid table-of-contents entries \cite{kumar2026stair}. On the SearchTome benchmark of 18 books across six domains, it reaches a Recall@1 of 82.6\%, compared with 76.9\% for a differentiable search index trained without the table of contents, and produces invalid section identifiers in only 0.05\% of predictions. Its gains over the structure-free model are largest for sections with few training examples, suggesting that exposing global document structure helps most where the model cannot learn that structure from data alone. STAIR, however, operates over static single-document hierarchies and supplies structure as prompt text, whereas CGA operates over a multi-document concept graph and supplies structure inside the attention computation. The two approaches therefore test complementary hypotheses about where structural information should enter a language model.

CGA builds on these developments but changes where retrieved knowledge enters the computation. Rather than using structure only to select or summarize textual evidence, CGA introduces concept-level structural information directly into transformer attention. The retrieved object is therefore not only a passage or summary, but a concept subgraph that can additionally carry temporal and epistemic qualification.

\subsection{Graph-Based RAG and Knowledge Graph Reasoning}

Graph-based RAG methods address the limitations of independent passage retrieval by explicitly representing entities, events, or semantic units and their relationships. GraphRAG constructs an entity graph from a document collection, detects graph communities, and generates community-level summaries that support global query-focused reasoning \cite{edge2024graphrag}. This approach is particularly effective for questions requiring an overview of an entire corpus. However, the graph is primarily used during indexing, retrieval, and summarization, while the language model ultimately receives textual summaries rather than the original graph topology.

Other approaches retrieve query-specific graph structures. GRAG retrieves relevant textual subgraphs and provides both textual and topological views to the generator \cite{hu2025grag}. GNN-RAG employs a graph neural network to identify relevant answer nodes and extracts shortest paths connecting question entities to candidate answers \cite{mavromatis2025gnnrag}. The resulting paths are verbalized and supplied to the language model as retrieval context. These methods improve multi-hop and multi-entity question answering, but graph reasoning remains a retrieval-stage operation whose output is converted into language-model-readable text.

LightRAG combines graph-enhanced indexing with dual-level retrieval over low-level entities and higher-level conceptual relationships \cite{guo2025lightrag}. Its incremental update mechanism also allows new information to be integrated without completely rebuilding the index. StructRAG takes a different approach by selecting an appropriate representation---such as a graph, table, catalogue, algorithm, or textual structure---according to the requirements of the query \cite{li2025structrag}. It then restructures the retrieved information and performs reasoning over the selected representation. These systems demonstrate that explicitly organized evidence can improve knowledge-intensive reasoning, but the constructed structures remain external inference contexts rather than persistent, temporally versioned objects integrated into the model's attention computation.

The broader integration of language models and knowledge graphs includes knowledge-graph-enhanced language models, language-model-assisted knowledge graph construction, and synergistic architectures combining neural and symbolic reasoning \cite{pan2024unifying}. Knowledge graph question-answering systems may use language models to generate graph queries, traverse relations, or verbalize reasoning paths, while graph neural networks and embedding models provide structured candidate retrieval. CGA is most closely related to the knowledge-graph-enhanced language-model direction, but differs by maintaining a dynamic graph of concepts rather than only entities or factual triples and by preserving temporal and epistemic metadata throughout attention-based reasoning.

\subsection{Graph Transformers and Knowledge-Enhanced Attention}

Graph transformers demonstrate that graph topology can be incorporated directly into attention rather than used only during retrieval. Graphormer augments self-attention with structural encodings derived from shortest-path distance, node centrality, and edge features \cite{ying2021graphormer}. Its results establish that graph-derived biases can effectively shape transformer computation. TokenGT adopts a different representation by treating graph nodes and edges as tokens and processing them through a largely standard transformer architecture \cite{kim2022tokengt}. This provides a flexible mechanism for graph learning but increases the number of input tokens with the size of the represented graph.

Earlier knowledge-enhanced language models also integrate symbolic knowledge into transformer representations. KnowBERT introduces entity linking and knowledge-attention modules that retrieve entity representations and recontextualize token representations using external knowledge bases \cite{peters2019knowbert}. K-BERT injects knowledge-graph triples into the input sequence and uses visibility constraints to control the propagation of the injected information \cite{liu2020kbert}. GreaseLM combines representations from a pretrained language model and a graph neural network through repeated cross-modal interaction layers for knowledge-based question answering \cite{zhang2022greaselm}.

These methods establish several mechanisms relevant to CGA: structural attention biases, graph tokenization, entity-aware attention, and repeated interaction between textual and graph representations. However, they generally assume static graphs or task-specific retrieved subgraphs and do not provide a lifecycle model for the represented knowledge. CGA combines graph-biased self-attention with cross-attention over retrieved concept nodes while additionally conditioning both mechanisms on temporal validity and epistemic status. The concept graph is therefore not only an auxiliary source of features, but a persistent and evolving knowledge representation queried at inference time.

\subsection{Temporal and Epistemic Knowledge Representation}

Knowledge graph embedding methods provide foundational representations for relational reasoning. TransE represents relations as translations between entity embeddings \cite{bordes2013transe}, while ComplEx and RotatE model richer relational patterns through complex-valued representations \cite{trouillon2016complex,sun2019rotate}. These approaches primarily operate over static graphs and do not explicitly model how facts or concepts change over time.

Temporal knowledge graph methods extend relational representations with timestamps and historical dependencies. TComplEx applies tensor decomposition to temporal knowledge-base completion, representing facts jointly across entity, relation, and temporal dimensions \cite{lacroix2020temporal}. RE-NET models temporal knowledge graphs as sequences of event structures and uses recurrent components to predict future events from historical graph states \cite{jin2020renet}. More recent work has explicitly incorporated uncertainty into temporal knowledge graph reasoning. DiffuTKG, for example, models future fact prediction as a diffusion process and introduces uncertainty regularization to reduce bias toward frequent events \cite{cai2024predicting}.

These approaches provide important foundations for temporal link prediction and event forecasting, but their objective is generally to predict missing or future graph facts. The CLM instead addresses the lifecycle of the represented conceptual units themselves. It distinguishes gradual change from discontinuous conceptual change and explicitly records concept birth, versioning, supersession, archiving, pruning, and invalidation. This formulation is informed by research on concept drift in data streams and diachronic semantic change \cite{gama2014survey,hamilton2016diachronic}, but applies these ideas to persistent graph-grounded concepts used during language-model reasoning.

Uncertainty has also been represented in probabilistic and uncertain knowledge graphs. UKGE learns embeddings for triples associated with confidence values \cite{chen2019ukge}, while large-scale knowledge-fusion systems estimate the probability of extracted claims using extraction confidence and source reliability \cite{dong2014knowledgevault}. Provenance standards such as PROV-O provide formal representations of the entities, agents, and activities involved in producing information \cite{w3c2013provo}. In the RAG literature, benchmarks such as RAGTruth demonstrate that supplying retrieved evidence does not eliminate unsupported or contradictory model outputs \cite{niu2024ragtruth}.

The epistemic model introduced in this work is complementary to probabilistic confidence estimation. Rather than assigning epistemic status solely from an extraction score, embedding score, or language-model probability, the CLM associates each graph relationship with the protocol through which it was established. Relationships may therefore be marked as hypothesised, validated through independent evidence, promoted after confirmation, or invalidated following contradiction. This allows temporal and provenance information to influence not only storage and retrieval, but also the weight assigned to a relationship during transformer attention.

Overall, existing research has separately demonstrated the value of external retrieval, graph-structured indexing, graph-aware attention, temporal knowledge graphs, provenance tracking, and uncertainty modelling. CGA brings these directions together by combining a persistent concept lifecycle model with structural self-attention and cross-attention over temporally scoped, epistemically weighted concept nodes. Its central distinction is therefore not merely the use of a knowledge graph, but the treatment of concepts as persistent computational objects whose structure, history, provenance, and epistemic status remain available during language-model reasoning.

\subsection{Attention Analysis and Mechanistic Interpretability}

A central question in this work is whether a change in attention implies a change in model behaviour. Prior work has shown that this relationship is not straightforward. Attention weights can often be substantially altered without changing model predictions, and gradient-based importance measures frequently disagree with attention-based ones \cite{jain2019attention,serrano2019attention}. Subsequent analysis argued that attention can still provide meaningful explanations under appropriate conditions and controls \cite{wiegreffe2019attention}. Mechanistic interpretability has since moved toward analysing the contributions of individual components to the residual stream \cite{elhage2021mathematical}, locating the computations responsible for specific predictions through causal interventions \cite{meng2022locating}, and steering model behaviour by adding directions directly to intermediate activations \cite{turner2023steering}.

These results motivate the evaluation design of the present study. CGA modifies attention by construction, so evidence that graph structure changes attention distributions is necessary but not sufficient to show that it changes what the model generates. We therefore pair every attention-level diagnostic with the same measurement under the disabled mechanism and with a controlled output-level comparison in which retrieval is held fixed.

%% file: 3-methodology.tex
\section{Proposed Methodology}\label{sec:methodology}

The proposed methodology consists of two closely connected components. The first is the CLM, which defines concepts as persistent, graph-grounded, and temporally versioned units of knowledge. The CLM specifies how concepts are created from evidence, how their representations and relationships evolve, and how they may be split, superseded, invalidated, pruned, or archived while preserving their historical states and provenance. Each concept is associated with a stable identity, a semantic representation, a relational neighbourhood, a version history, and an epistemic status describing the confirmation protocol supporting its relationships.

The second component is CGA, a neuro-symbolic transformer architecture that uses the CLM as an external knowledge substrate. CGA retrieves a query-relevant and temporally scoped concept subgraph and introduces it directly into transformer computation through graph-biased self-attention and cross-attention over concept-node representations. The complete methodology further includes concept extraction, concept-graph construction, query-to-concept alignment, multi-path retrieval, evidence-path adjudication, and joint optimization of the language-model and graph-attention components.

\subsection{Problem Formulation}
\label{subsec:problem-formulation}

Let $\mathcal{D}=\{d_1,d_2,\ldots,d_N\}$ denote a corpus of source documents and let $\mathcal{G}_\tau=(\mathcal{C}_\tau,\mathcal{E}_\tau)$ denote the concept graph available to the system at transaction time $\tau$. The set $\mathcal{C}_\tau$ contains persistent concept nodes extracted or inferred from $\mathcal{D}$, while $\mathcal{E}_\tau$ contains typed relationships between these concepts. Each concept and relationship may possess temporal validity information, provenance references, lifecycle status, and epistemic status.

Given a natural-language query $q$, the system first identifies an optional temporal anchor $t_q$ and a permission context $P_q$. The temporal anchor specifies the point in time for which the requested knowledge should be resolved. When no temporal expression is detected, $t_q$ is assigned the current valid-time state. The permission context determines the subset of concepts and relationships that may be accessed by the querying user or application.

The objective is to generate a response
\begin{equation}
    R = F(q,\mathcal{D},\mathcal{G}_\tau,t_q,P_q),
\end{equation}
where $F$ denotes the complete CGA retrieval and synthesis process. The generated response should satisfy four principal requirements:

\begin{enumerate}
    \item \textbf{Factual grounding.}
    Every factual claim in $R$ should be supported by accessible evidence from $\mathcal{D}$ or by relationships in $\mathcal{G}_\tau$ whose provenance can be traced to that evidence. Claims based only on inferred graph structure should be explicitly distinguished from independently validated claims.

    \item \textbf{Temporal correctness.}
    When a temporal anchor $t_q$ is provided, the response should reflect the concepts and relationships that were valid at $t_q$, rather than automatically returning their most recent states. Concepts that had not yet emerged at $t_q$ should not contribute to the answer, while archived or invalidated concepts may remain available when they were valid during the requested period.

    \item \textbf{Structural faithfulness.}
    The response should preserve the relevant relationships contained in the retrieved concept subgraph and support reasoning over multi-hop paths that may connect evidence distributed across several documents. The generated answer should not introduce graph relationships that are unsupported by either direct evidence or explicitly identified structural inference.

    \item \textbf{Epistemic calibration.}
    The response should communicate the epistemic status of the knowledge on which it relies. Validated and promoted relationships may contribute at full weight, hypothesised relationships should contribute at reduced weight and be presented with appropriate uncertainty, and invalidated relationships should be excluded from current-state reasoning while remaining available for historically anchored queries.
\end{enumerate}

To operationalize these requirements, the system retrieves an accessible, query-relevant, and temporally valid subgraph
\begin{equation}
    \mathcal{G}_{q,t,P}
    =
    \operatorname{Retrieve}
    \left(
        q,
        \mathcal{G}_\tau,
        t_q,
        P_q
    \right),
\end{equation}
subject to
\begin{equation}
    \mathcal{C}_{q,t,P}
    \subseteq
    \left\{
        c\in\mathcal{C}_\tau
        \mid
        \operatorname{valid}(c,t_q)=1
        \land
        \operatorname{accessible}(c,P_q)=1
    \right\}.
\end{equation}

An analogous constraint is applied to the retrieved relationships. The resulting subgraph is supplied to the dual-attention mechanism together with the query representation and the supporting textual evidence. The methodological problem addressed by CGA can therefore be stated as follows: determine whether directly incorporating a temporally resolved and epistemically weighted concept subgraph into transformer attention improves factual, structural, temporal, and epistemic reasoning compared with systems that use only flat textual retrieval or serialize graph structures into textual context.

The framework is specified in full, but the empirical study in Section~\ref{sec:evaluation} evaluates it in parts: the structural components (Forms A and B) on multi-hop question answering with retrieval held fixed, point-in-time resolution on a temporal benchmark, and protocol-derived epistemic status on a controlled synthetic graph. The adjudicator, the temporal and epistemic training objectives, and the full lifecycle dynamics are not evaluated; their detailed formulations are given in the appendix.

\subsection{Concept Lifecycle Model}
\label{subsec:clm}

The CLM defines the representational and temporal principles governing the concepts stored in the knowledge graph. Unlike conventional knowledge graphs in which nodes are treated as static entities, the CLM models concepts as persistent computational objects whose semantic representation, relational context, evidential support, and epistemic status may change over time. A concept therefore maintains a stable identity while preserving the history of the states through which it has evolved.

\subsubsection{Formal Concept Representation}

A concept $C_i$ is represented as the tuple
\begin{equation}
    C_i =
    \left(
        id_i,
        \mathbf{z}_i,
        \mathcal{N}_i,
        b_i,
        \mathcal{V}_i,
        \Pi_i,
        s_i
    \right),
\end{equation}
where:

\begin{itemize}
    \item $id_i$ is a persistent identifier that remains unchanged throughout the lifecycle of the concept;
    \item $\mathbf{z}_i \in \mathbb{R}^{d_c}$ is the current semantic representation of the concept;
    \item $\mathcal{N}_i$ is the set of typed and weighted relationships connecting $C_i$ to other concepts;
    \item $b_i$ is an immutable birth record containing the creation time, creation mechanism, and founding evidence;
    \item $\mathcal{V}_i$ is an ordered sequence of historical concept states;
    \item $\Pi_i$ maps each relationship and supporting claim to its provenance and epistemic status;
    \item $s_i$ is the current lifecycle status of the concept.
\end{itemize}

The semantic representation $\mathbf{z}_i$ is computed from the evidence associated with the concept. Let
\begin{equation}
    \mathcal{X}_i =
    \left\{
        \mathbf{x}_j
        \mid
        x_j \text{ supports } C_i
    \right\}
\end{equation}
denote the set of embeddings of the supporting text segments. A basic prototype representation is given by
\begin{equation}
    \mathbf{z}_i
    =
    \frac{1}{|\mathcal{X}_i|}
    \sum_{\mathbf{x}_j \in \mathcal{X}_i}
    \mathbf{x}_j.
\end{equation}

This centroid may be replaced by a learned aggregation function in the implemented system. The concept meaning is not determined by $\mathbf{z}_i$ alone, however. It is jointly determined by the semantic representation and the relational neighbourhood $\mathcal{N}_i$. Two concepts with similar embeddings may therefore remain distinct when they occupy different structural, temporal, or domain-specific positions in the graph.

The version history is defined as
\begin{equation}
    \mathcal{V}_i
    =
    \left[
        v_i^{(1)},
        v_i^{(2)},
        \ldots,
        v_i^{(m)}
    \right],
\end{equation}
where each version contains
\begin{equation}
    v_i^{(k)}
    =
    \left(
        t_{\mathrm{valid}}^{(k)},
        t_{\mathrm{transaction}}^{(k)},
        \mathbf{z}_i^{(k)},
        \mathcal{N}_i^{(k)},
        s_i^{(k)}
    \right).
\end{equation}

Here, $t_{\mathrm{valid}}$ denotes the period in which a state is considered valid in the represented domain, while $t_{\mathrm{transaction}}$ records when that state was introduced into the system. This distinction allows the graph to represent delayed evidence, retrospective corrections, and historically valid claims that were recorded only at a later time.

For a query anchored at time $t_q$, the applicable version is selected by
\begin{equation}
    \operatorname{Resolve}(C_i,t_q)
    =
    \underset{v \in \mathcal{V}_i}{\arg\max}
    \left\{
        t_{\mathrm{valid}}(v)
        \mid
        t_{\mathrm{valid}}(v) \leq t_q
    \right\}.
\end{equation}

If no version satisfies the temporal condition, the concept is treated as unavailable for that query because it had not yet emerged or was not valid at the requested time.

\subsubsection{Concept Emergence}

Concepts are created from evidence rather than assumed to exist in advance. The CLM defines four complementary emergence mechanisms.

\paragraph{Embedding-density emergence.}
A candidate concept may be created when a sufficiently coherent group of evidence embeddings forms a dense region in the representation space. Let $\boldsymbol{\mu}_k$ denote the centroid of a candidate cluster. A concept candidate is produced when
\begin{equation}
    \left|
        \left\{
            \mathbf{x}_j :
            \cos(\mathbf{x}_j,\boldsymbol{\mu}_k)
            \geq \rho
        \right\}
    \right|
    \geq \delta_D,
\end{equation}
where $\rho$ is the minimum similarity to the centroid and $\delta_D$ is the minimum density required for concept creation.

\paragraph{Extraction-based emergence.}
A concept may also be created when the concept-extraction model repeatedly identifies the same semantic category across independent evidence segments. Given a candidate label $\ell$, creation is triggered when
\begin{equation}
    \left|
        \left\{
            x_j :
            p(\ell \mid x_j) \geq \tau_C
        \right\}
    \right|
    \geq \delta_C,
\end{equation}
where $\tau_C$ is the minimum extraction confidence and $\delta_C$ is the minimum number of supporting segments.

\paragraph{Hierarchical splitting.}
An existing concept may become semantically heterogeneous as new evidence accumulates. When its evidence embeddings form stable subclusters with sufficient internal cohesion and intercluster separation, the concept may be divided into more specific child concepts. The parent concept remains in the graph, while the new concepts are linked through typed \texttt{parent\_of} relationships.

\paragraph{Graph-predicted emergence.}
A graph model may predict that an unobserved concept would explain a set of otherwise disconnected or weakly connected relationships. Concepts produced in this manner are introduced as hypotheses rather than validated knowledge. They receive reduced influence during attention and must accumulate direct evidence before promotion.

Each concept birth is recorded through
\begin{equation}
    b_i =
    \left(
        \kappa_i,
        t_i,
        \mathcal{S}_i^{0}
    \right),
\end{equation}
where $\kappa_i$ identifies the emergence mechanism, $t_i$ is the creation time, and $\mathcal{S}_i^{0}$ is the founding evidence set.

\subsubsection{Concept Evolution}

The CLM distinguishes between gradual semantic evolution and discontinuous conceptual change. Let
\begin{equation}
    \Delta_i^{(k)}
    =
    1-
    \cos
    \left(
        \mathbf{z}_i^{(k)},
        \mathbf{z}_i^{(k-1)}
    \right)
\end{equation}
denote the semantic displacement between consecutive versions.

Gradual evolution occurs when the semantic representation or neighbourhood changes incrementally without altering the fundamental identity of the concept. In this case, a new version is appended to $\mathcal{V}_i$, while $id_i$ remains unchanged:
\begin{equation}
    C_i^{(k-1)}
    \longrightarrow
    C_i^{(k)}.
\end{equation}

Discontinuous change occurs when the semantic displacement and its rate of change exceed predefined or learned thresholds:
\begin{equation}
    \Delta_i^{(k)} > \delta_{\mathrm{shift}}
    \quad \land \quad
    \frac{\Delta_i^{(k)}-\Delta_i^{(k-1)}}
         {t_k-t_{k-1}}
    > v_{\mathrm{shift}}.
\end{equation}

In this case, the new state is treated as a distinct concept $C_j$ with a new persistent identifier:
\begin{equation}
    C_j
    \xrightarrow{\texttt{evolved\_from}}
    C_i.
\end{equation}

The original concept is archived rather than deleted, allowing queries anchored before the transition to resolve to the historically appropriate state. The distinction between gradual and discontinuous change is evaluated empirically because suitable thresholds may vary across domains and concept types.

\subsubsection{Epistemic Status}

Every relationship $e_{ij}\in\mathcal{E}_\tau$ is associated with an epistemic status
\begin{equation}
    \pi(e_{ij})
    \in
    \{
        \textit{hypothesised},
        \textit{validated},
        \textit{promoted},
        \textit{invalidated}
    \}.
\end{equation}

A \textit{hypothesised} relationship is inferred from semantic similarity, graph prediction, or insufficiently confirmed extraction evidence. It may contribute to reasoning at reduced weight but should not be presented as independently established knowledge.

A \textit{validated} relationship is supported through an accepted confirmation protocol. Such a protocol may include agreement across independent source families, direct structured evidence, or human verification. Independence is assessed using observable factors such as authorship, institutional origin, publication pipeline, temporal separation, and duplicate or citation lineage.

A \textit{promoted} relationship was initially hypothesised but subsequently accumulated sufficient confirming evidence. It is then permitted to contribute at the same operational weight as a validated relationship, while retaining its promotion history for auditability.

An \textit{invalidated} relationship has been contradicted or withdrawn. Its weight is set to zero for current-state queries, but the relationship is preserved together with its original validity interval so that historical queries remain accurate.

The attention-scale coefficient associated with an edge is therefore
\begin{equation}
    \alpha(e_{ij})
    =
    \begin{cases}
        1, & \pi(e_{ij}) \in
        \{\textit{validated},\textit{promoted}\},\\[4pt]
        h_{ij}, & \pi(e_{ij})=\textit{hypothesised},\\[4pt]
        0, & \pi(e_{ij})=\textit{invalidated},
    \end{cases}
\end{equation}
where $0<h_{ij}<1$ is the reduced hypothesis weight. For historically anchored queries, an invalidated edge may recover its pre-invalidation weight when the requested time precedes the invalidation event.

\subsubsection{Archiving, Invalidation, and Pruning}

The lifecycle status of a concept is defined as
\begin{equation}
    s_i
    \in
    \{
        \textit{active},
        \textit{archived},
        \textit{invalidated},
        \textit{pruned}
    \}.
\end{equation}

An \textit{active} concept continues to receive evidence and may acquire new versions or relationships. An \textit{archived} concept is no longer updated because it has been superseded or has become inactive, but it remains available for historical resolution. An \textit{invalidated} concept is retained together with the evidence and timestamp responsible for its invalidation. A \textit{pruned} concept is a candidate that failed to accumulate sufficient support within a persistence window.

Pruning differs from invalidation. Invalidation applies to concepts or relationships that were previously accepted but were later contradicted. Pruning applies to candidates that never acquired sufficient evidence to become operationally active. In both cases, the associated provenance is retained to support auditing, reproducibility, and possible reactivation when new evidence becomes available.

\subsection{Concept Graph Construction}
\label{subsec:graph-construction}

The Concept Graph Construction process instantiates the CLM from an evolving document corpus. It transforms heterogeneous textual evidence into persistent concept nodes, typed relationships, temporal states, and provenance links. The resulting graph is not constructed once and treated as fixed; it is updated incrementally as new evidence arrives and as existing concepts change their semantic or epistemic status.

\subsubsection{Document Processing and Concept Extraction}

Each document $d_n\in\mathcal{D}$ is first divided into semantically coherent evidence units:
\begin{equation}
    d_n
    \longrightarrow
    \mathcal{U}_n
    =
    \left\{
        u_{n1},
        u_{n2},
        \ldots,
        u_{nm_n}
    \right\}.
\end{equation}

The segmentation process respects sentence, paragraph, clause, and discourse boundaries so that each unit expresses one or a small number of related propositions. For each evidence unit $u_j$, the system computes an embedding
\begin{equation}
    \mathbf{x}_j
    =
    f_{\mathrm{enc}}(u_j)
    \in \mathbb{R}^{d_e},
\end{equation}
and applies a concept-extraction model
\begin{equation}
    \mathcal{A}_j
    =
    f_{\mathrm{ext}}(u_j),
\end{equation}
where $\mathcal{A}_j$ contains candidate concept mentions, semantic labels, relation candidates, temporal expressions, and extraction confidence scores.

Each extracted item is linked to a provenance record
\begin{equation}
    p_j
    =
    \left(
        d_n,
        u_j,
        a_n,
        o_n,
        t_n,
        \ell_n
    \right),
\end{equation}
where $a_n$ denotes authorship, $o_n$ the institutional or source origin, $t_n$ the source timestamp, and $\ell_n$ the source-lineage metadata. These attributes are later used when estimating evidence independence and assigning epistemic status.

\subsubsection{Concept Identification and Deduplication}

For each extracted candidate $a_j$, the system determines whether it should be assigned to an existing concept or create a new concept. Candidate-to-concept compatibility is computed using semantic, structural, lexical, and provenance features:
\begin{equation}
    S(a_j,C_i)
    =
    \lambda_{\mathrm{sem}}
    S_{\mathrm{sem}}(a_j,C_i)
    +
    \lambda_{\mathrm{lex}}
    S_{\mathrm{lex}}(a_j,C_i)
    +
    \lambda_{\mathrm{str}}
    S_{\mathrm{str}}(a_j,C_i)
    +
    \lambda_{\mathrm{temp}}
    S_{\mathrm{temp}}(a_j,C_i),
\end{equation}
where the coefficients are either tuned on a development set or learned jointly.

The semantic component may be defined as
\begin{equation}
    S_{\mathrm{sem}}(a_j,C_i)
    =
    \cos
    \left(
        \mathbf{x}_j,
        \mathbf{z}_i
    \right).
\end{equation}

The lexical component measures agreement between extracted labels and concept aliases. The structural component compares the candidate relationships with the existing neighbourhood of $C_i$, while the temporal component penalizes assignments that conflict with known concept-validity intervals.

The candidate is assigned to
\begin{equation}
    i^{*}
    =
    \underset{i}{\arg\max}\,
    S(a_j,C_i)
\end{equation}
when
\begin{equation}
    S(a_j,C_{i^{*}})
    \geq
    \tau_{\mathrm{assign}}.
\end{equation}

Otherwise, the candidate enters the concept-emergence process described in Section~\ref{subsec:clm}. This separation prevents every extracted phrase from immediately becoming a graph node.

Potential duplicates are detected through high semantic similarity, overlapping aliases, comparable neighbourhoods, and shared evidence. However, two candidates are not merged solely because their embeddings are similar. A merge is permitted only when the available semantic, structural, temporal, and provenance evidence indicates that they refer to the same persistent concept. Merge operations are recorded so that the original concept identifiers and supporting evidence remain traceable.

\subsubsection{Relationship Construction}

The graph contains typed relationships
\begin{equation}
    e_{ij}
    =
    \left(
        C_i,
        r_{ij},
        C_j,
        w_{ij},
        \pi_{ij},
        T_{ij},
        P_{ij}
    \right),
\end{equation}
where $r_{ij}$ is the relation type, $w_{ij}$ its base strength, $\pi_{ij}$ its epistemic status, $T_{ij}$ its temporal annotation, and $P_{ij}$ its provenance set.

The canonical relation vocabulary includes:

\begin{itemize}
    \item \texttt{related\_to}, representing general semantic association;
    \item \texttt{co\_occurs\_with}, representing repeated direct co-occurrence;
    \item \texttt{parent\_of}, representing hierarchical specialization;
    \item \texttt{evolved\_from}, representing discontinuous lifecycle transitions;
    \item \texttt{subject\_to}, representing dependency on a rule, condition, or constraint;
    \item \texttt{triggered\_by}, representing causal or event-based dependence;
    \item \texttt{contradicts}, representing explicit semantic or evidential conflict;
    \item \texttt{equivalent\_to}, representing semantic equivalence across contexts.
\end{itemize}

A relation candidate may be produced by direct extraction, repeated co-occurrence, semantic similarity, graph prediction, or human annotation. Its initial edge weight is computed from the available evidence:
\begin{equation}
    w_{ij}
    =
    \beta_{\mathrm{ext}}
    c_{\mathrm{ext}}
    +
    \beta_{\mathrm{freq}}
    c_{\mathrm{freq}}
    +
    \beta_{\mathrm{sem}}
    c_{\mathrm{sem}}
    +
    \beta_{\mathrm{src}}
    c_{\mathrm{src}},
\end{equation}
where $c_{\mathrm{ext}}$ is the extraction confidence, $c_{\mathrm{freq}}$ the normalized occurrence frequency, $c_{\mathrm{sem}}$ the semantic compatibility score, and $c_{\mathrm{src}}$ the source-support score.

The operational edge contribution used by downstream reasoning is
\begin{equation}
    \widetilde{w}_{ij}
    =
    \alpha(e_{ij})w_{ij},
\end{equation}
where $\alpha(e_{ij})$ is determined by the epistemic status defined in the CLM. This ensures that a high graph-prediction score does not automatically give a hypothesised relationship the same influence as independently validated evidence.

\subsubsection{Temporal and Epistemic Annotation}

Each node and relationship carries a bitemporal record that separates valid time, when an item is considered true in the represented domain, from transaction time, when the system recorded or modified it. When contradicting evidence arrives, the current validity interval is closed and an invalidation event is created, while the previous state remains queryable for historical anchors. Epistemic status is assigned by a configurable confirmation rule based on the number of independent source families supporting a relationship, human confirmation, or confirmation from an accepted structured source. The full formal specification is given in Appendix~\ref{app:temporal-epistemic-annotation}. Simplified forms of the valid-time record and of the confirmation rule are evaluated in Sections~\ref{subsec:temporal-results} and~\ref{subsec:epistemic-results}.

\subsubsection{Concept--Evidence Incidence Matrix}

To preserve the association between concept nodes and their supporting evidence, the system maintains a concept--evidence incidence matrix
\begin{equation}
    \mathbf{W}
    \in
    \mathbb{R}^{|\mathcal{C}|\times|\mathcal{U}|},
\end{equation}
where $\mathcal{U}$ is the set of evidence units.

Each element $W_{ij}$ represents the strength of association between concept $C_i$ and evidence unit $u_j$:
\begin{equation}
    W_{ij}
    =
    \lambda_{\mathrm{cos}}
    \cos(\mathbf{z}_i,\mathbf{x}_j)
    +
    \lambda_{\mathrm{conf}}
    c_{ij}^{\mathrm{ext}}
    +
    \lambda_{\mathrm{freq}}
    c_{ij}^{\mathrm{freq}}
    +
    \lambda_{\mathrm{prov}}
    c_{ij}^{\mathrm{prov}}.
\end{equation}

Here, $c_{ij}^{\mathrm{ext}}$ is the extraction confidence, $c_{ij}^{\mathrm{freq}}$ is the normalized evidence frequency, and $c_{ij}^{\mathrm{prov}}$ reflects the quality and independence of the supporting provenance.

The matrix has three roles. First, it enables retrieval of the textual evidence supporting a concept. Second, it provides the evidence-weighting signal used by concept-node cross-attention. Third, it maintains traceability from generated answers to source documents.

As new evidence arrives, $\mathbf{W}$ is updated incrementally. Concept prototypes, graph relationships, and epistemic states are recomputed only for the affected neighbourhood rather than rebuilding the complete graph. The output of this process is the temporally versioned and epistemically annotated graph
\begin{equation}
    \mathcal{G}_\tau
    =
    \left(
        \mathcal{C}_\tau,
        \mathcal{E}_\tau,
        \mathbf{W}_\tau
    \right),
\end{equation}
which serves as the structured knowledge substrate for retrieval, adjudication, and Concept-Grounded Attention.

\subsection{Concept-Grounded Attention}
\label{subsec:cga}

CGA integrates the query-relevant concept graph directly into transformer computation. The architecture combines two complementary mechanisms. The first, referred to as \emph{Form A}, introduces graph-derived structural information as an additive bias within self-attention. The second, referred to as \emph{Form B}, performs cross-attention from the language-model representations to the retrieved concept nodes. Form A modifies how textual tokens attend to one another according to the relationships among their associated concepts, whereas Form B allows the model to access concept-level information that may not be explicitly present in the input text.

Let the tokenized input query and supporting textual context contain $n$ tokens. At transformer layer $\ell$, their hidden representations are denoted by
\begin{equation}
    \mathbf{H}^{(\ell)}
    \in
    \mathbb{R}^{n\times d},
\end{equation}
where $d$ is the model hidden dimension. Let
\begin{equation}
    \mathcal{G}_{q,t,P}
    =
    \left(
        \mathcal{C}_{q,t,P},
        \mathcal{E}_{q,t,P}
    \right)
\end{equation}
denote the query-relevant, temporally valid, and permission-compatible concept subgraph produced by the retrieval procedure. The number of retrieved concept nodes is bounded by
\begin{equation}
    k =
    |\mathcal{C}_{q,t,P}|
    \leq K_{\max}.
\end{equation}

\subsubsection{Query-to-Concept Alignment}

Before applying graph-aware attention, the system maps relevant token spans to concept nodes. A concept-linking model first identifies candidate concept mentions in the query and supporting context:
\begin{equation}
    \mathcal{A}_q
    =
    f_{\mathrm{link}}
    \left(
        q,
        \mathcal{C}_{q,t,P}
    \right).
\end{equation}

A span-alignment component then produces the mapping
\begin{equation}
    M:
    \{1,\ldots,n\}
    \rightarrow
    \mathcal{C}_{q,t,P}
    \cup
    \{\varnothing\},
\end{equation}
where
\begin{equation}
    M(i)=C_a
\end{equation}
indicates that token position $i$ belongs to a span linked to concept $C_a$, while
\begin{equation}
    M(i)=\varnothing
\end{equation}
indicates that no concept association was identified.

The mapping is accompanied by a confidence value
\begin{equation}
    \mu_i
    =
    p\left(
        M(i)=C_a
        \mid
        q,
        \mathcal{G}_{q,t,P}
    \right).
\end{equation}

Tokens that are not linked to a concept receive no graph-derived attention bias. This provides graceful degradation for novel terminology, extraction failures, and queries referring to concepts that are not yet represented in the graph.

\subsubsection{Form A: Graph-Biased Self-Attention}

For a transformer attention head, the conventional query, key, and value matrices are
\begin{equation}
    \mathbf{Q}
    =
    \mathbf{H}^{(\ell)}
    \mathbf{W}_{Q},
    \qquad
    \mathbf{K}
    =
    \mathbf{H}^{(\ell)}
    \mathbf{W}_{K},
    \qquad
    \mathbf{V}
    =
    \mathbf{H}^{(\ell)}
    \mathbf{W}_{V},
\end{equation}
where
\begin{equation}
    \mathbf{Q},
    \mathbf{K},
    \mathbf{V}
    \in
    \mathbb{R}^{n\times d_h}
\end{equation}
and $d_h$ is the attention-head dimension.

Form A introduces a graph-derived bias matrix
\begin{equation}
    \mathbf{B}_{G}
    \in
    \mathbb{R}^{n\times n}.
\end{equation}

For token positions $i$ and $j$, the corresponding bias is defined as
\begin{equation}
B_G[i,j]
=
\begin{cases}
\begin{aligned}
\mu_i\mu_j \Bigl(
    &\theta_1 \widetilde{w}_{ab}
    + \theta_2 \phi_{\mathrm{path}}(C_a,C_b)
    \\[-2pt]
    &+ \theta_3 \phi_{\mathrm{central}}(C_a,C_b)
    + \theta_4 \phi_{\mathrm{time}}(C_a,C_b,t_q)
\Bigr),
\end{aligned}
&
\substack{
M(i)=C_a,\\
M(j)=C_b,
}
\\[12pt]
0,
&
M(i)=\varnothing
\ \lor\
M(j)=\varnothing.
\end{cases}
\end{equation}

Here, $\widetilde{w}_{ab}$ is the epistemically adjusted edge strength:
\begin{equation}
    \widetilde{w}_{ab}
    =
    \alpha(e_{ab})w_{ab},
\end{equation}
where $w_{ab}$ is the base relation weight and $\alpha(e_{ab})$ is determined by the epistemic status of the relationship.

The path component may be defined from the graph distance:
\begin{equation}
    \phi_{\mathrm{path}}(C_a,C_b)
    =
    \begin{cases}
        \exp
        \left(
            -\gamma_{\mathrm{path}}
            d_{\mathcal{G}}(C_a,C_b)
        \right),
        &
        d_{\mathcal{G}}(C_a,C_b)<\infty,
        \\[4pt]
        0,
        &
        \text{otherwise},
    \end{cases}
\end{equation}
where $d_{\mathcal{G}}$ is the shortest-path distance in the retrieved subgraph.

The centrality component summarizes the structural prominence of the two nodes:
\begin{equation}
    \phi_{\mathrm{central}}(C_a,C_b)
    =
    \frac{1}{2}
    \left(
        \operatorname{cent}(C_a)
        +
        \operatorname{cent}(C_b)
    \right),
\end{equation}
where $\operatorname{cent}$ may be instantiated using normalized degree, PageRank, or another graph-centrality measure.

The temporal component reduces the influence of relationships whose valid-time state is distant from the query anchor:
\begin{equation}
    \phi_{\mathrm{time}}(C_a,C_b,t_q)
    =
    \exp
    \left(
        -\gamma_t
        \Delta_t(e_{ab},t_q)
    \right),
\end{equation}
where $\Delta_t(e_{ab},t_q)$ measures the temporal distance between the relationship state and the query anchor.

Graph-biased self-attention is then computed as
\begin{equation}
    \mathbf{H}_{A}
    =
    \operatorname{softmax}
    \left(
        \frac{\mathbf{Q}\mathbf{K}^{\top}}
             {\sqrt{d_h}}
        +
        \mathbf{B}_{G}
    \right)
    \mathbf{V}.
\end{equation}

The coefficients
\begin{equation}
    \boldsymbol{\theta}
    =
    \left[
        \theta_1,
        \theta_2,
        \theta_3,
        \theta_4
    \right]
\end{equation}
are learned during training. Near-zero initialization is used so that the initial behavior remains close to that of the pretrained transformer before graph-aware fine-tuning.

\subsubsection{Form B: Cross-Attention over Concept Nodes}

Form B gives the language model direct access to the representations of the retrieved concept nodes. For each concept
\begin{equation}
    C_j
    \in
    \mathcal{C}_{q,t,P},
\end{equation}
a concept representation is constructed by combining its semantic embedding, structural encoding, temporal state, epistemic representation, and query relevance:
\begin{equation}
    \mathbf{c}_j
    =
    f_{\mathrm{concept}}
    \left(
        \mathbf{z}_j,
        \mathbf{g}_j,
        \mathbf{t}_j,
        \mathbf{p}_j,
        r_j
    \right).
\end{equation}

Here:

\begin{itemize}
    \item $\mathbf{z}_j$ is the semantic concept embedding;
    \item $\mathbf{g}_j$ is a precomputed neighbourhood or graph encoding;
    \item $\mathbf{t}_j$ represents temporal state and recency;
    \item $\mathbf{p}_j$ represents lifecycle and epistemic status;
    \item $r_j$ is the query-relevance score assigned during retrieval.
\end{itemize}

The concept-node matrix is
\begin{equation}
    \mathbf{C}_q
    =
    \begin{bmatrix}
        \mathbf{c}_1^\top\\
        \mathbf{c}_2^\top\\
        \vdots\\
        \mathbf{c}_k^\top
    \end{bmatrix}
    \in
    \mathbb{R}^{k\times d_c}.
\end{equation}

The concept keys and values are produced through learned projections:
\begin{equation}
    \mathbf{K}_c
    =
    \mathbf{C}_q
    \mathbf{P}_{K},
    \qquad
    \mathbf{V}_c
    =
    \mathbf{C}_q
    \mathbf{P}_{V},
\end{equation}
where
\begin{equation}
    \mathbf{K}_c,
    \mathbf{V}_c
    \in
    \mathbb{R}^{k\times d_h}.
\end{equation}

Each concept node is assigned an operational gate
\begin{equation}
    g_j
    =
    r_j
    \cdot
    a_j
    \cdot
    \alpha_j,
\end{equation}
where $r_j$ is query relevance, $a_j$ indicates permission and temporal availability, and $\alpha_j$ is the epistemic coefficient associated with the node and its supporting relationships. The gate vector is
\begin{equation}
    \mathbf{g}
    =
    \left[
        g_1,\ldots,g_k
    \right]^\top.
\end{equation}

The concept-attention score matrix is
\begin{equation}
    \mathbf{S}_{B}
    =
    \frac{
        \mathbf{Q}_{A}
        \mathbf{K}_c^\top
    }{
        \sqrt{d_h}
    }
    +
    \mathbf{1}_n
    \log
    \left(
        \mathbf{g}
        +
        \epsilon
    \right)^\top,
\end{equation}
where $\mathbf{Q}_A$ is a learned projection of $\mathbf{H}_A$, $\mathbf{1}_n$ is an $n$-dimensional vector of ones, and $\epsilon>0$ ensures numerical stability.

The additive logarithmic gate preserves dimensional consistency and operates as a prior over concept-node attention. Concept nodes with low epistemic or retrieval weight receive reduced probability, while unavailable or invalidated nodes can be masked by assigning their gate a value approaching zero.

Form B is computed as
\begin{equation}
    \mathbf{H}_{B}
    =
    \operatorname{softmax}
    \left(
        \mathbf{S}_{B}
    \right)
    \mathbf{V}_c.
\end{equation}

\subsubsection{Dual-Attention Composition}

The outputs of Form A and Form B are combined through a learned gate:
\begin{equation}
    \boldsymbol{\eta}
    =
    \sigma
    \left(
        \left[
            \mathbf{H}_{A}
            \Vert
            \mathbf{H}_{B}
        \right]
        \mathbf{W}_{\eta}
        +
        \mathbf{b}_{\eta}
    \right),
\end{equation}
where $\Vert$ denotes concatenation and
\begin{equation}
    \boldsymbol{\eta}
    \in
    [0,1]^{n\times d_h}.
\end{equation}

The combined representation is
\begin{equation}
    \mathbf{H}_{\mathrm{CGA}}
    =
    \mathbf{H}_{A}
    +
    \boldsymbol{\eta}
    \odot
    \mathbf{H}_{B},
\end{equation}
where $\odot$ denotes element-wise multiplication.

The combined representation is added to the residual stream:
\begin{equation}
    \widetilde{\mathbf{H}}^{(\ell)}
    =
    \mathbf{H}^{(\ell)}
    +
    \mathbf{H}_{\mathrm{CGA}}.
\end{equation}

In pre-normalisation transformers, which normalise the input of each sublayer rather than the residual stream, no additional normalisation should be applied at this point. Normalising $\mathbf{H}^{(\ell)}+\mathbf{H}_{\mathrm{CGA}}$ rescales the representation received by all subsequent layers and alters the pretrained computation even when $\boldsymbol{\eta}=\mathbf{0}$. Together with a closed initial gate ($\mathbf{b}_{\eta}\ll0$), the formulation above leaves the pretrained computation essentially unchanged at initialisation. Section~\ref{subsec:form-b-analysis} shows that violating this condition was sufficient to collapse generation.

The output is passed to the transformer feed-forward sublayer in the standard manner. The dual-attention layer may be inserted into every transformer block or into a selected subset of layers. The latter option reduces computational overhead and is treated as an implementation hyperparameter evaluated during ablation.

The two forms serve distinct purposes. Form A enhances structural relationships among concepts already expressed in the input, whereas Form B introduces knowledge contained in the retrieved graph but absent from the textual sequence. Their combined use allows both the textual evidence and the persistent concept graph to influence generation.

\subsection{Retrieval and Adjudication}
\label{subsec:retrieval-adjudication}

The retrieval process identifies the subset of the concept graph required to answer a query. Because no single retrieval strategy is optimal for all query types, CGA uses three complementary paths: vector similarity retrieval, graph traversal, and temporally scoped retrieval. Their outputs are evaluated by an adjudication component that selects or combines the retrieved evidence.

Let
\begin{equation}
    \mathbf{q}
    =
    f_{\mathrm{enc}}(q)
\end{equation}
denote the query embedding. The retrieval objective is to construct
\begin{equation}
    \mathcal{G}_{q,t,P}
    =
    \operatorname{Adjudicate}
    \left(
        \mathcal{R}_A,
        \mathcal{R}_B,
        \mathcal{R}_C
    \right),
\end{equation}
where $\mathcal{R}_A$, $\mathcal{R}_B$, and $\mathcal{R}_C$ are the outputs of the three retrieval paths.

\subsubsection{Path A: Semantic Retrieval}

Path A retrieves concept nodes according to semantic similarity between the query and the concept representations:
\begin{equation}
    \mathcal{R}_A
    =
    \operatorname{TopK}_{C_i\in\mathcal{C}_\tau}
    \cos
    \left(
        \mathbf{q},
        \mathbf{z}_i
    \right).
\end{equation}

The retrieval is constrained by temporal and permission filters:
\begin{equation}
    \operatorname{valid}(C_i,t_q)=1,
    \qquad
    \operatorname{accessible}(C_i,P_q)=1.
\end{equation}

Path A provides fast and high-recall retrieval but does not independently capture multi-hop topology. It is therefore suitable for direct or single-concept queries and serves as the initial retrieval path in the sequential strategy.

The representative embedding of the retrieved set may be computed as
\begin{equation}
    \overline{\mathbf{r}}_A
    =
    \frac{
        \sum_{C_i\in\mathcal{R}_A}
        s_i^{A}\mathbf{z}_i
    }{
        \sum_{C_i\in\mathcal{R}_A}
        s_i^{A}
    },
\end{equation}
where $s_i^{A}$ is the normalized semantic-retrieval score.

\subsubsection{Path B: Multi-Hop Graph Retrieval}

Path B starts from seed concepts identified through query-to-concept alignment:
\begin{equation}
    \mathcal{S}_q
    =
    \left\{
        M(i)
        \mid
        M(i)\neq\varnothing
    \right\}.
\end{equation}

From these seeds, the system traverses the graph up to a maximum hop depth $H_{\max}$. The score of a path
\begin{equation}
    \pi
    =
    \left(
        C_{i_0},
        e_{i_0i_1},
        \ldots,
        e_{i_{h-1}i_h},
        C_{i_h}
    \right)
\end{equation}
is defined as
\begin{equation}
    S_{\mathrm{path}}(\pi)
    =
    \left[
        \prod_{r=0}^{h-1}
        \widetilde{w}_{i_ri_{r+1}}
    \right]
    \exp(-\gamma_h h)
    S_{\mathrm{rel}}(C_{i_h},q),
\end{equation}
where $\widetilde{w}$ contains epistemic weighting, $h$ is the path length, and $S_{\mathrm{rel}}$ measures the relevance of the terminal concept to the query.

The graph-retrieval output is
\begin{equation}
    \mathcal{R}_B
    =
    \operatorname{TopPaths}
    \left(
        \mathcal{S}_q,
        H_{\max},
        K_B
    \right).
\end{equation}

Only temporally valid and permission-compatible nodes and edges are traversed. Invalidated edges receive zero weight for current-state queries, while hypothesised edges may remain traversable at reduced weight.

A representative embedding for Path B is obtained from the retrieved subgraph:
\begin{equation}
    \overline{\mathbf{r}}_B
    =
    \sum_{C_i\in\mathcal{R}_B}
    s_i^{B}\mathbf{z}_i,
\end{equation}
where the weights $s_i^{B}$ are derived from path relevance, centrality, and epistemic strength.

\subsubsection{Path C and Adjudication}

A third retrieval path, Path C, resolves each candidate concept and relationship to the state valid at the query anchor $t_q$, distinguishing valid time from transaction time so that evidence valid only after $t_q$ is excluded even when it is semantically similar to the query. An adjudicator then scores the outputs of Paths A, B, and C using semantic relevance, concept coverage, epistemic quality, temporal compatibility, and estimated cost, and either selects a single path or fuses several. Retrieval proceeds sequentially, executing the more expensive paths only when semantic retrieval is judged insufficient, and a rule-based policy is used until sufficient routing supervision is available. The complete formulation of Path C, adjudicator scoring, sequential early stopping, and cold-start routing is given in Appendix~\ref{app:routing}. The multi-hop evaluation uses frozen outputs of the semantic and graph retrieval paths; a simplified form of point-in-time resolution is evaluated in Section~\ref{subsec:temporal-results}, and adjudication is not evaluated.

\subsection{Training Objective and Complexity}
\label{subsec:training-complexity}

The proposed architecture is trained in stages to preserve the capabilities of the pretrained language model while progressively introducing concept-aware reasoning. The concept-extraction, linking, graph-retrieval, adjudication, and dual-attention components may first be trained independently and subsequently optimized jointly.

\subsubsection{Initialization and Training Stages}

The training procedure consists of three principal stages.

\paragraph{Stage 1: Component pretraining.}

The concept-extraction and linking components are trained using annotated or automatically derived concept mentions, relation labels, and span alignments. The retrieval and adjudication components are trained using query--evidence pairs and known optimal retrieval paths.

The base language model remains unchanged during this stage.

\paragraph{Stage 2: Dual-attention insertion.}

The Form A and Form B components are inserted into selected transformer layers. The graph-bias coefficients and dual-attention gate are initialized near zero:
\begin{equation}
    \theta_r
    \sim
    \mathcal{N}
    \left(
        0,
        \sigma_{\theta}^{2}
    \right),
    \qquad
    \sigma_{\theta}\ll 1,
\end{equation}
and
\begin{equation}
    \mathbf{b}_{\eta}
    \ll 0,
\end{equation}
so that the initial cross-attention contribution remains small.

The projection matrices
\begin{equation}
    \mathbf{P}_{K},
    \mathbf{P}_{V},
    \mathbf{W}_{\eta}
\end{equation}
are initialized using Xavier initialization.

This initialization preserves the pretrained model's behavior at the beginning of fine-tuning and reduces instability caused by randomly initialized graph components.

\paragraph{Stage 3: Joint fine-tuning.}

The complete architecture is jointly trained on knowledge-intensive question answering, temporal reasoning, and graph-grounded synthesis tasks. The training data include both positive examples and controlled negative examples involving irrelevant nodes, invalid temporal states, hypothesised relationships, and corrupted graph paths.

\subsubsection{Language-Modelling Objective}

The primary generation loss is the autoregressive language-model loss:
\begin{equation}
    \mathcal{L}_{\mathrm{LM}}
    =
    -
    \sum_{r=1}^{T}
    \log
    p
    \left(
        y_r
        \mid
        y_{<r},
        q,
        \mathcal{G}_{q,t,P}
    \right),
\end{equation}
where
\begin{equation}
    y_1,\ldots,y_T
\end{equation}
are the ground-truth response tokens.

\subsubsection{Concept-Linking Objective}

For each token position $i$, the concept-linking component predicts the target concept assignment $m_i^{*}$. Its loss is
\begin{equation}
    \mathcal{L}_{\mathrm{link}}
    =
    -
    \sum_{i=1}^{n}
    \log
    p
    \left(
        M(i)=m_i^{*}
        \mid
        q
    \right).
\end{equation}

When only span-level labels are available, this objective can be decomposed into concept-detection and span-alignment losses.

\subsubsection{Structural-Attention Objective}

The Form A auxiliary objective encourages graph-biased attention to align with relevant concept relationships. Let
\begin{equation}
    Y_{ij}^{A}
    \in
    \{0,1\}
\end{equation}
indicate whether the concepts associated with tokens $i$ and $j$ participate in a relevant supporting relation or reasoning path.

Let
\begin{equation}
    A_{ij}^{G}
\end{equation}
denote the normalized graph-aware attention assigned between the positions. The structural loss may be defined as
\begin{equation}
    \mathcal{L}_{A}
    =
    -
    \sum_{i,j}
    \left[
        Y_{ij}^{A}
        \log A_{ij}^{G}
        +
        \left(
            1-Y_{ij}^{A}
        \right)
        \log
        \left(
            1-A_{ij}^{G}
        \right)
    \right].
\end{equation}

This objective is applied only to token pairs for which concept alignment is available. Positive labels may be derived from supporting-fact annotations, gold graph paths, or high-confidence extracted relationships.

\subsubsection{Concept-Retrieval Objective}

Form B is trained to assign greater attention to concepts supporting the target answer. Let
\begin{equation}
    y_j^{B}
    \in
    \{0,1\}
\end{equation}
indicate whether concept $C_j$ belongs to the gold supporting subgraph, and let $\widehat{r}_j$ be its predicted relevance.

A binary relevance loss is
\begin{equation}
    \mathcal{L}_{B}
    =
    -
    \sum_{j=1}^{k}
    \left[
        y_j^{B}
        \log \widehat{r}_j
        +
        \left(
            1-y_j^{B}
        \right)
        \log
        \left(
            1-\widehat{r}_j
        \right)
    \right].
\end{equation}

Alternatively, when a ranking of relevant concepts is available, a pairwise ranking objective may be used:
\begin{equation}
    \mathcal{L}_{\mathrm{rank}}
    =
    \sum_{j^{+},j^{-}}
    \max
    \left(
        0,
        m
        -
        s_{j^{+}}
        +
        s_{j^{-}}
    \right),
\end{equation}
where $j^{+}$ and $j^{-}$ denote relevant and irrelevant concept nodes.

\subsubsection{Temporal, Epistemic, and Routing Objectives}

When the corresponding supervision is available, training can additionally include an epistemic calibration objective $\mathcal{L}_{\mathrm{epi}}$ with an invalidation penalty $\mathcal{L}_{\mathrm{invalid}}$, a temporal objective $\mathcal{L}_{\mathrm{temp}}$ that penalises reliance on states not valid at $t_q$, and adjudicator objectives $\mathcal{L}_{\mathrm{route}}$ and $\mathcal{L}_{\mathrm{cost}}$ for path selection and retrieval cost. Their definitions are given in Appendix~\ref{app:objectives}. These objectives were not used in the experiments reported here, which did not include temporal, epistemic, or routing supervision.

\subsubsection{Combined Objective}

The complete training objective is
\begin{align}
    \mathcal{L}_{\mathrm{total}}
    ={}&
    \mathcal{L}_{\mathrm{LM}}
    +
    \lambda_{\mathrm{link}}
    \mathcal{L}_{\mathrm{link}}
    +
    \lambda_A
    \mathcal{L}_A
    +
    \lambda_B
    \mathcal{L}_B
    \nonumber\\
    &+
    \lambda_{\mathrm{epi}}
    \mathcal{L}_{\mathrm{epi}}
    +
    \lambda_{\mathrm{invalid}}
    \mathcal{L}_{\mathrm{invalid}}
    +
    \lambda_{\mathrm{temp}}
    \mathcal{L}_{\mathrm{temp}}
    \nonumber\\
    &+
    \lambda_{\mathrm{route}}
    \mathcal{L}_{\mathrm{route}}
    +
    \lambda_{\mathrm{cost}}
    \mathcal{L}_{\mathrm{cost}}.
\end{align}

The loss weights are selected using a validation set. Not every training instance requires every objective. For example, $\mathcal{L}_{\mathrm{temp}}$ is applied only when temporal supervision is available, while $\mathcal{L}_{\mathrm{epi}}$ requires epistemically labelled graph evidence.

\subsubsection{Computational Complexity}

Let:

\begin{itemize}
    \item $n$ denote the textual sequence length;
    \item $d$ denote the hidden dimension;
    \item $d_h$ denote the attention-head dimension;
    \item $k\leq K_{\max}$ denote the number of retrieved concept nodes;
    \item $m\leq n$ denote the number of tokens mapped to concept nodes;
    \item $|\mathcal{C}|$ and $|\mathcal{E}|$ denote the numbers of concept nodes and graph edges.
\end{itemize}

The standard transformer self-attention cost per layer is
\begin{equation}
    O(n^2d_h).
\end{equation}

Form A retains this dominant complexity. Constructing a dense graph-bias matrix has worst-case cost
\begin{equation}
    O(m^2),
\end{equation}
although it can be reduced by computing biases only for mapped token pairs connected within a bounded graph distance. Under sparse computation, its cost becomes proportional to the number of relevant mapped pairs.

Form B computes attention between $n$ textual positions and $k$ concept nodes:
\begin{equation}
    O(nkd_h).
\end{equation}

Because $k$ is bounded by $K_{\max}$, Form B grows linearly with the number of retrieved concepts rather than with the complete graph size.

The additional memory required for the concept keys and values per layer is
\begin{equation}
    O(kd_h),
\end{equation}
while the cross-attention score matrix requires
\begin{equation}
    O(nk)
\end{equation}
memory unless memory-efficient attention is used.

Vector retrieval has approximate cost
\begin{equation}
    O(\log|\mathcal{C}|)
\end{equation}
under an approximate nearest-neighbour index, subject to the properties of the selected indexing method.

Bounded graph traversal has worst-case cost
\begin{equation}
    O(b^{H_{\max}}),
\end{equation}
where $b$ is the effective branching factor. In practice, pruning by relation type, epistemic weight, query relevance, and path score limits the explored neighbourhood.

Temporal resolution over an indexed version history has cost
\begin{equation}
    O(\log|\mathcal{V}_i|)
\end{equation}
per concept when version timestamps are stored in ordered form.

The concept--evidence incidence matrix is sparse because each evidence unit supports only a small subset of concepts. Its storage cost is therefore
\begin{equation}
    O(\operatorname{nnz}(\mathbf{W}))
\end{equation}
rather than
\begin{equation}
    O(|\mathcal{C}||\mathcal{U}|),
\end{equation}
where $\operatorname{nnz}$ denotes the number of non-zero entries.

The dominant inference cost remains standard self-attention when
\begin{equation}
    k \ll n.
\end{equation}
For short queries with large retrieved subgraphs, Form B may instead become significant. The limits $K_{\max}$, $H_{\max}$, and the number of transformer layers containing CGA are therefore treated as primary efficiency hyperparameters.

The sequential adjudication procedure further reduces average retrieval cost by avoiding graph traversal and temporal resolution for queries that can be answered reliably through semantic retrieval alone. The empirical evaluation measures this trade-off through latency, memory use, path-activation frequency, and answer quality.

%% file: 4-evaluation.tex
\section{Experimental Setup}
\label{sec:evaluation}

The evaluation has three parts. The first tests the structural components of CGA on multi-hop question answering: concept-graph retrieval, graph-biased self-attention (Form A), and cross-attention over concept nodes (Form B). To separate the internal effect of the attention mechanisms from the effect of retrieval, all compared systems receive identical, frozen retrieval outputs, and every attention diagnostic is compared with the same measurement taken with the mechanism disabled. The second part tests CLM point-in-time resolution on a public temporal benchmark, and the third tests protocol-derived epistemic status on a controlled synthetic graph.

\subsection{Research Questions}
\label{subsec:research-questions}

\paragraph{RQ1: Does concept-graph retrieval recover multi-hop structure unavailable to passage retrieval?}
We measure whether the graph-enhanced pipeline returns explicit concept paths connecting distributed evidence, and whether it improves evidence recall.

\paragraph{RQ2: Does Form A change the model's attention and outputs?}
We compare the attention and the answers of Form A with those of an identical configuration in which Form A is disabled, and we measure how large the graph bias would need to be to change the model's decisions.

\paragraph{RQ3: Why did Form B fail, and can it be repaired?}
We isolate the components of the initial Form B implementation responsible for its failure and evaluate an identity-preserving variant.

\paragraph{RQ4: Does explicit temporal resolution improve answers to time-dependent questions?}
We compare CLM point-in-time resolution with retrieval that ignores time, retrieval that serialises dates, and a recency heuristic.

\paragraph{RQ5: Does protocol-derived epistemic status reduce unsupported assertions?}
We compare models given raw reports, claims with their sources, and claims with epistemic status derived by the CLM confirmation protocol.

\subsection{Multi-Hop Data}
\label{subsec:datasets}

The multi-hop evaluation uses MuSiQue \cite{trivedi2022musique} and HotpotQA \cite{yang2018hotpotqa}. Deterministic, disjoint manifests contain 500 training, 100 validation, and 100 test questions per dataset, selected as the first questions of each annotation stratum. The test set comprises 200 questions in four strata of 50: MuSiQue two-hop, MuSiQue three-hop, HotpotQA bridge, and HotpotQA comparison. For each split, a corpus-level concept graph is built from all passages of that split (Section~\ref{subsec:graph-construction}), using spaCy named entities, noun chunks, and noun keyphrases as concept mentions and co-occurrence and embedding similarity as relations. The resulting test graphs contain 21{,}497 (MuSiQue) and 20{,}105 (HotpotQA) concepts. Gold supporting facts are used only to derive gold concept paths and to label concepts as gold-linked or distractor-linked for evaluation.

\subsection{Compared Systems}
\label{subsec:baselines}

\paragraph{Closed book and dense RAG.}
The closed-book model answers from the question alone. Dense RAG supplies the top five passages by embedding similarity \cite{lewis2020retrieval}.

\paragraph{Hybrid graph retrieval.}
The hybrid control fuses dense retrieval with multi-hop graph traversal (Paths A and B in Section~\ref{subsec:retrieval-adjudication}) by reciprocal-rank fusion and supplies the top five passages to the unmodified model. It is the principal control for Forms A and B.

\paragraph{Serialised concept graph.}
This baseline supplies the same passages together with the retrieved concept paths serialised as text, within the same total input budget.

\paragraph{CGA-Form A.}
Form A receives exactly the same inputs as the hybrid control and adds the graph bias of Section~\ref{subsec:cga} to self-attention in encoder layers 4 and 5. Because the multi-hop data contain no temporal anchors or epistemic distinctions, all relations are treated as validated ($\alpha=1$), and the bias uses the edge-strength and path-distance terms with learned coefficients $\theta_1$ and $\theta_2$.

\paragraph{CGA-Form B and Form A+B.}
Form B applies gated cross-attention from the encoder hidden states to the retrieved concept nodes after encoder blocks 4 and 5, alone or combined with Form A.

\paragraph{Controls.}
Every Form A diagnostic is repeated with Form A disabled ($\theta=0$). In addition, the bias is scaled by factors of $10$ to $10^4$, and two corrupted variants shuffle token positions within the bias matrix (random graph) or permute token-to-concept links (shuffled links). For Form B, the initial module is evaluated untrained, with its gate forced to zero, and trained, and is compared with an identity-preserving variant (Section~\ref{subsec:form-b-analysis}). To separate concept information from added trainable capacity, the identity-preserving Form B is also trained and tested with \emph{mismatched concepts} taken from another question, for one and three epochs, and a query-conditioned variant of Form A is compared with a control whose relevance features are shuffled (Section~\ref{subsec:capacity-controls}).

\subsection{Multi-Hop Metrics and Diagnostics}
\label{subsec:evaluation-metrics}

Answers are scored by exact match (EM) and token F1 after standard normalisation. Retrieval is scored by passage Recall@5 and by exact and soft graph-path recall: exact path recall credits a gold path only when a retrieved path matches it completely, and soft path recall credits the fraction of its concepts contained in the best-matching retrieved path.

For Form A we record, per question, the mean attention mass (averaged over heads and the two modified layers) on token pairs whose concepts are both gold-linked and on all other concept-linked pairs, and their ratio. Token pairs within the same concept are included in the main measure; we additionally report a strict variant that excludes them and a distance-matched variant that restricts distractor pairs to the positional distance range of the gold pairs. To quantify the attention--output gap we record the ratio between the bias magnitude and the standard deviation of the pre-softmax attention logits, the minimum margin between the two most probable decoder tokens along the baseline answer, the maximum change in decoder logits induced by Form A under teacher forcing, and the relative change of the encoder output. For Form B we record the gate activation, the attention mass per gold and per distractor concept, the relative change of the hidden state, and the mean pairwise cosine similarity of the projected concept keys.

\subsection{Temporal Evaluation}
\label{subsec:temporal-setup}

Temporal resolution is evaluated on the knowledge-update (78) and temporal-reasoning (133) questions of LongMemEval, using the oracle sessions \cite{wu2025longmemeval}. Each knowledge-update question is accompanied by two dated sessions, the later of which updates a fact stated in the earlier one; temporal-reasoning questions require ordering or relating dated events. Evidence units are user turns, each carrying its session date as valid time. For each question the same four most similar units are retrieved for all systems, and a fixed generator (flan-t5-large, 512 input tokens, greedy decoding) produces the answer. The systems differ only in how time is represented: \emph{dense} presents the units without dates; \emph{dense, dated} presents them chronologically with dates; \emph{recency only} keeps only units from the latest session; and \emph{CLM resolution} groups units into concepts by embedding similarity ($\cos\geq0.5$), orders the versions of each concept by valid time, marks the latest as current and earlier ones as superseded, and keeps unrelated units as distinct dated events. Answers are scored by whether the normalised gold answer is contained in the prediction.

\subsection{Epistemic Evaluation}
\label{subsec:epistemic-setup}

Epistemic status is evaluated on a synthetic graph with fictional entities, so that answers cannot come from pretraining. Each instance concerns one relation of a fictional company, reported on dated days by news outlets that belong to four publisher families; outlets of the same family have unrelated names, and family membership is stated only in an ownership registry. The CLM confirmation rule of Section~\ref{subsec:clm} with $n_{\mathrm{indep}}=2$ assigns status: validated when a claim is reported by two independent families, hypothesised when reported only by outlets of one family, invalidated when retracted, and promoted when a hypothesis is later confirmed by a second family. Five categories are balanced: a validated claim (expected output: the entity), a hypothesised claim reported by two outlets of the same family (``unconfirmed: entity''), a validated claim with a competing hypothesis (the validated entity), a retracted claim (``unknown''), and a promoted claim (the entity). Training and test sets contain 2{,}000 and 500 instances with disjoint entity names. Three input formats carry the same information: raw reports with the ownership registry, serialised claims with their sources but no status, and serialised claims with protocol-derived status, from which invalidated claims are removed. For each format, flan-t5-small is fine-tuned for three epochs. Outputs are scored by their epistemic decision (assert, unconfirmed, or unknown) and the entity they select, and we report the unsupported-assertion rate: presenting a hypothesised claim as fact, relying on a retracted claim, or preferring an unconfirmed competitor.

\subsection{Larger-Model Evaluation}
\label{subsec:api-setup}

The temporal and epistemic experiments require only text generation, so we repeated them with three much larger instruction-tuned models served through an OpenAI-compatible inference endpoint: \texttt{qwen2.5-72b}, \texttt{gpt-oss-120b}, and \texttt{qwen35-122b}, a 122-billion-parameter Qwen model (identifiers as served by the endpoint). All generation used temperature 0. The temporal systems received exactly the same retrieved units and prompts as before. For the epistemic task these models were not fine-tuned; instead, every input format was preceded by the same instruction stating the confirmation rule (two independent publisher groups are required, outlets of one group count as one source, retracted claims must not be used, and graph statuses map to established or not established) and the three permitted output forms. The test therefore asks whether a large model can apply the rule itself from raw reports or from claims with their sources, or needs the protocol-derived status. Because large models paraphrase, temporal answers from all generators, including flan-t5-large, were additionally scored by an LLM judge (gpt-oss-120b) using the LongMemEval judging prompts for knowledge-update and temporal-reasoning questions \cite{wu2025longmemeval}. The epistemic run for gpt-oss-120b was executed twice; the two runs differed by at most 2.4 points in any aggregate measure.

\subsection{Implementation and Statistical Analysis}
\label{subsec:implementation-details}
\label{subsec:statistical-analysis}

The multi-hop experiments use google/flan-t5-small (revision \texttt{0fc9ddf}) with all-MiniLM-L6-v2 embeddings, a 384-token input budget, at most 12 generated tokens, and greedy decoding, run in float32 on the CPU of an Apple M1 Max. The base model is frozen. Form A was trained for three epochs (1{,}500 updates) with learning rate $10^{-5}$, initial $\theta=0.01$, and structural loss weight $\lambda_A=0.1$, selected by a staged search on validation data in which all candidates tied. The initial Form B was trained jointly with Form A on 60 examples for one epoch (learning rate $5\times10^{-4}$, $\lambda_A=\lambda_B=0.1$, where $\mathcal{L}_B$ is the negative log attention mass on gold concepts). Repaired Form B variants were trained without Form A, with the same loss and learning rate, on 60 or 500 examples for one epoch. Latency was measured after 20 warm-up queries over 200 questions and three repetitions.

Trainable configurations use seeds 13, 42, and 77. Because Form A trains only two deterministic scalars on a frozen model, its three seeds are identical, and its statistics are computed over the 200 questions. For all other comparisons, seed outcomes are averaged within question before testing. Paired differences are reported with 10{,}000-sample bootstrap 95\% confidence intervals and two-sided Wilcoxon signed-rank tests. Data manifests and per-question outputs are available from the authors on reasonable request.

%% file: 5-results.tex
\section{Results}
\label{sec:results}

This section reports the multi-hop evaluation of the structural components (Sections~\ref{subsec:answer-level}--\ref{subsec:efficiency-results}), followed by the evaluation of temporal resolution (Section~\ref{subsec:temporal-results}) and epistemic status (Section~\ref{subsec:epistemic-results}). Unless stated otherwise, multi-hop results refer to the fixed set of 200 test questions, and statistical tests are computed per question.

\subsection{Answer-Level Results}
\label{subsec:answer-level}

Table~\ref{tab:answer-level} reports answer quality for all multi-hop configurations. Absolute scores are low for every system, reflecting the small generator and the 384-token input budget. Retrieval improved over the closed-book model, and hybrid graph retrieval improved over dense retrieval (F1 0.188 versus 0.164). Serialising the retrieved concept paths as text did not improve over hybrid retrieval (F1 0.178). Form A produced exactly the same answers as the hybrid control. The initial Form B implementation, alone or combined with Form A, collapsed answer generation (F1 0.006); Section~\ref{subsec:form-b-analysis} identifies the cause.

\begin{table}[t]
\centering
\small
\caption{Answer-level results on the 200 multi-hop test questions (flan-t5-small, greedy decoding). Passage Recall@5 refers to the retrieved evidence}
\label{tab:answer-level}
\begin{tabular}{lccc}
\hline
\textbf{System} & \textbf{EM} & \textbf{F1} & \textbf{Passage R@5} \\
\hline
Closed book & 0.100 & 0.145 & -- \\
Dense RAG & 0.110 & 0.164 & 0.687 \\
Hybrid graph retrieval & 0.130 & 0.188 & 0.662 \\
Serialised concept graph & 0.135 & 0.178 & 0.662 \\
CGA-Form A & 0.130 & 0.188 & 0.662 \\
CGA-Form B (initial) & 0.000 & 0.006 & 0.662 \\
CGA-Form A+B (initial) & 0.000 & 0.006 & 0.662 \\
\hline
\end{tabular}
\end{table}

\subsection{Structural Evidence Retrieval}
\label{subsec:structural-retrieval}

The graph-enhanced pipeline returns explicit paths through the concept graph, which dense retrieval cannot represent. It achieved a soft path recall of 0.244 and an exact path recall of 0.015, compared with zero for dense retrieval. Soft path recall was 0.310 on MuSiQue two-hop, 0.112 on MuSiQue three-hop, 0.269 on HotpotQA bridge, and 0.282 on HotpotQA comparison questions, indicating that path ranking becomes harder as the number of required transitions grows.

Graph retrieval did not, however, improve evidence recall. Hybrid passage Recall@5 (0.662) was slightly below dense Recall@5 (0.687). Graph traversal recovered a passage missed by dense retrieval for 6 of 200 questions, but introduced at least one additional non-supporting passage for 170. The answer-level gain of hybrid over dense retrieval therefore cannot be attributed to better recall of supporting passages.

\subsection{Form A: Attention Diagnostics and the Attention--Output Gap}
\label{subsec:graph-sensitive-attention}

\paragraph{Diagnostics with Form A enabled.}
With Form A enabled, gold-linked token pairs received a mean graph-associated attention mass of 0.008125 and distractor-linked pairs 0.002959, a mean gold-to-distractor ratio of 2.758 (median 2.46), a ratio above one for all 197 questions with both pair types, and identical answers to the hybrid control (F1 0.1877). The three training seeds produced identical coefficients ($\theta_1=0.00999$, $\theta_2=0.00761$), because only these two deterministic scalars were trained on a frozen model; counting the three seeds separately would give 591 question--seed observations, but these correspond to 197 independent questions, over which all statistics are computed.

\begin{figure}[t]
    \centering
    \includegraphics[width=\linewidth]{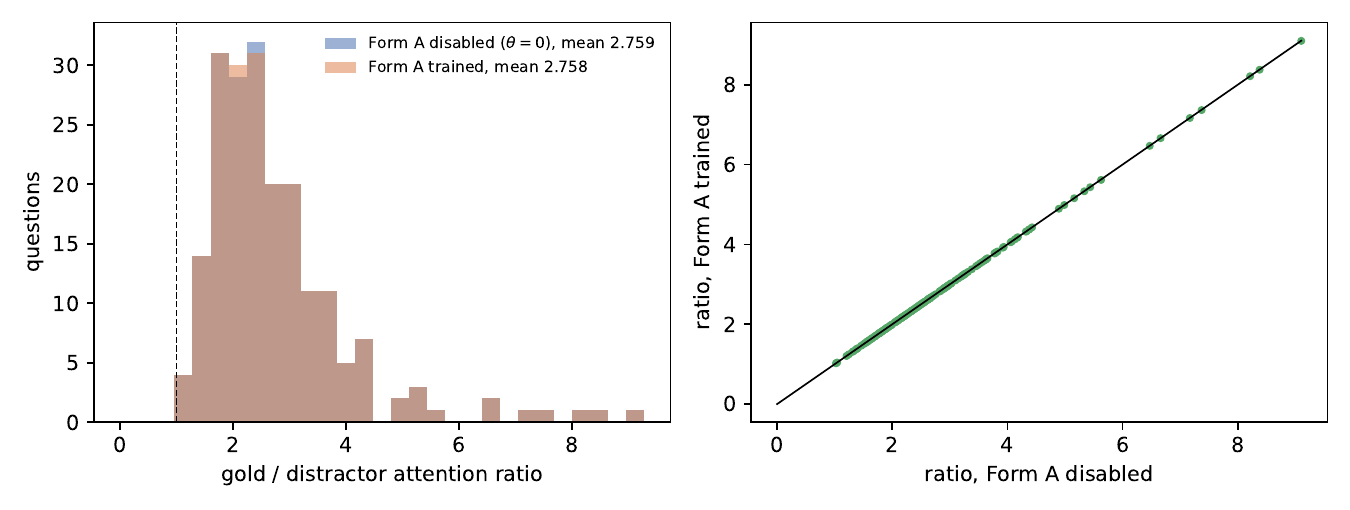}
    \caption{Gold-to-distractor attention ratio for 197 test questions with Form A disabled ($\theta=0$) and with the trained Form A coefficients. Left: distributions. Right: per-question comparison; all points lie on the diagonal. The apparent evidence of graph-steered attention is reproduced exactly without Form A}
    \label{fig:theta0-control}
\end{figure}

\paragraph{The attention ratio is a property of the pretrained model.}
These diagnostics suggest that Form A steers attention toward gold concepts. Disabling Form A ($\theta=0$) while keeping every other input identical yielded a mean ratio of 2.759, compared with 2.758 with the trained coefficients (Figure~\ref{fig:theta0-control}). The per-question difference never exceeded 0.011 in absolute value; its mean was $-0.0012$ (95\% CI $[-0.0014,-0.0010]$), a small but systematic \emph{decrease}. The gold-versus-distractor preference is therefore already present in the attention of the pretrained flan-t5-small encoder and is not produced by Form A. The preference is partly positional: excluding token pairs within the same concept lowered the ratio to 2.29, and additionally restricting distractor pairs to the distance range of gold pairs lowered it to 2.03.

\paragraph{Why Form A has no effect.}
Form A does modify the computation, but at a negligible scale. The mean absolute bias on mapped token pairs was 0.0020, whereas the pre-softmax attention logits at the modified layers had a standard deviation of 4.30, a ratio of $4.6\times10^{-4}$. The bias assigned to gold pairs exceeded that assigned to distractor pairs (0.0025 versus 0.0018), but this difference follows from graph structure rather than learning: gold concepts co-occur in supporting sentences and are therefore connected by stronger edges. At the output, Form A changed decoder logits by a median maximum of 0.006, whereas the median minimum margin between the two most probable tokens of the baseline answer was 0.93. The change exceeded the margin for one question in 200, and no greedy decision changed.

\paragraph{Scaling the bias.}
Table~\ref{tab:theta-sweep} scales the trained coefficients by factors of 10 to $10^4$. Answers began to change at $10\times$ and $100\times$ (2 and 9 questions) without a significant effect on F1. At $10^3\times$ and $10^4\times$, where the bias is comparable to the attention logits, 122 and 136 answers changed and F1 decreased (0.157 and 0.111; the latter significantly, $p=0.003$). Stronger graph bias of this form therefore disrupts rather than improves generation, and the gold-to-distractor ratio did not increase at any scale.

\begin{table}[t]
\centering
\small
\caption{Effect of scaling the trained Form A coefficients. $|\Delta\ell|$: median maximum change in decoder logits; $\rho$: relative change of the encoder output; changed: answers differing from $\theta=0$}
\label{tab:theta-sweep}
\begin{tabular}{rcccccc}
\hline
\textbf{Scale} & \textbf{Mean $|B_G|$} & \textbf{$|\Delta\ell|$} & \textbf{$\rho$} & \textbf{Changed} & \textbf{F1} & \textbf{Ratio} \\
\hline
0 & 0 & 0 & 0 & -- & 0.188 & 2.759 \\
1 & 0.002 & 0.006 & 0.0003 & 0 & 0.188 & 2.758 \\
10 & 0.020 & 0.063 & 0.003 & 2 & 0.188 & 2.748 \\
100 & 0.196 & 0.654 & 0.033 & 9 & 0.190 & 2.648 \\
1{,}000 & 1.957 & 7.38 & 0.338 & 122 & 0.157 & 2.257 \\
10{,}000 & 19.57 & 10.01 & 0.424 & 136 & 0.111 & 2.445 \\
\hline
\end{tabular}
\end{table}

\subsection{Form B: Failure Diagnosis and Repair}
\label{subsec:form-b-analysis}

The initial Form B implementation collapsed generation (Table~\ref{tab:answer-level}). Table~\ref{tab:form-b} isolates the cause. The initial module applied a freshly initialised layer normalisation to the sum of the encoder hidden state and the gated concept signal. flan-t5-small is a pre-normalisation transformer whose residual stream carries unnormalised hidden states, so this operation rescales the representation passed to all subsequent layers. Before any training, the initial module changed the encoder hidden states by 97\% of their norm and reduced F1 to 0.003, with 41 empty answers. Forcing the gate to zero left this result unchanged, which shows that the normalisation alone, not the concept signal, caused the collapse. Training on 60 examples did not recover the model (F1 0.005).

An identity-preserving variant removes the extra normalisation, computes $\mathbf{H}+\boldsymbol{\eta}\odot\mathbf{H}_B$, and initialises the gate closed (bias $-4$). Untrained, it reproduced the baseline answers exactly. Trained on 60 examples it changed one answer on average; trained on 500 examples it raised F1 from 0.188 to 0.195 in all three seeds (paired difference $+0.008$, 95\% CI $[-0.002,0.021]$, $p=0.35$), changing 27 answers on average. Restricting cross-attention to the eight most relevant concepts with stronger gold-concept supervision gave a similar result ($+0.006$, CI $[0.000,0.014]$, $p=0.11$). Neither improvement is statistically significant, and Section~\ref{subsec:capacity-controls} shows that the one-epoch gain does not depend on the concepts at all.

\begin{table}[t]
\centering
\small
\caption{Form B diagnosis and repair (mean over seeds 13, 42, 77; untrained rows are seed-independent in F1). $\rho$: relative change of the hidden state at the Form B layer. Gold/distr.: mean Form B attention per gold and per distractor concept}
\label{tab:form-b}
\begin{tabular}{lcccccc}
\hline
\textbf{Variant} & \textbf{Train} & \textbf{F1} & \textbf{$\Delta$F1} & \textbf{$\rho$} & \textbf{Gate} & \textbf{Gold/distr.} \\
\hline
No Form B & -- & 0.188 & -- & -- & -- & -- \\
Initial & 0 & 0.003 & $-0.185$ & 0.968 & 0.50 & 1.15 \\
Initial, $\eta=0$ & 0 & 0.003 & $-0.185$ & 0.968 & 0.00 & 1.15 \\
Initial & 60 & 0.005 & $-0.183$ & 0.968 & 0.62 & 2.47 \\
Identity-preserving & 0 & 0.188 & 0.000 & 0.000 & 0.02 & 1.19 \\
Identity-preserving & 60 & 0.188 & 0.000 & 0.000 & 0.34 & 2.53 \\
Identity-preserving & 500 & 0.195 & $+0.008$ & 0.0003 & 0.53 & 3.51 \\
\quad + top-8, $\lambda_B=1$ & 500 & 0.194 & $+0.006$ & 0.0002 & 0.53 & 3.60 \\
\hline
\end{tabular}
\end{table}

The attention diagnostics also show that Form B did not attend mainly to distractors, as its total attention mass would suggest. Because gold concepts made up only 24\% of the retrieved concepts, a majority of total attention mass on distractors is compatible with a preference for gold concepts. Per concept, the trained initial module already attended 2.5 times more to gold than to distractor concepts, and the repaired module trained on 500 examples 3.5 times more. Form B therefore learned to favour gold concepts, and its collapse was architectural rather than a failure of relevance learning. Section~\ref{subsec:capacity-controls} shows, however, that this preference is driven by the concept features rather than by the question. The mean pairwise cosine similarity of the projected concept keys increased with training (0.37 untrained, 0.65 after 500 examples), indicating some convergence of concept representations, but not collapse.

\subsection{Controls for Added Trainable Capacity}
\label{subsec:capacity-controls}

A trained Form B adds parameters to a frozen model, and any improvement could therefore reflect added capacity rather than concept information. We repeated the training of the identity-preserving Form B with \emph{mismatched concepts}, giving every question the concept nodes, relevance scores, and gold labels of a different question in both training and testing. We also trained both variants for three epochs instead of one. Finally, we evaluated a query-conditioned variant of Form A designed to address the diagnosis of Section~\ref{subsec:graph-sensitive-attention}. It adds a learned key-side bias $\lambda\,\mu_j\,\sigma(w_1 r_j + w_2 g_j + b)$ towards tokens whose concepts are relevant to the question, where $r_j$ is the embedding similarity between the question and the concept of token $j$, $g_j$ its graph proximity to the question's concepts, and $\lambda$ is initialised at 1. It is trained with gold-concept supervision and compared with a control in which the per-token features are shuffled.

Table~\ref{tab:capacity} summarises the results. With one epoch, mismatched concepts reproduced the gain of real concepts exactly, with 198 of 200 identical answers. With three epochs, Form B with real concepts improved F1 significantly over the model without Form B ($+0.033$, 95\% CI $[0.007,0.062]$, $p=0.024$), but mismatched concepts achieved $+0.025$. The concept-specific contribution, real minus mismatched, was $+0.008$ (CI $[0.000,0.020]$, $p=0.09$). Per-concept attention averaged over the test set was identical for real and mismatched concepts, which indicates that Form B's attention depended on the concept features rather than on the question. The query-conditioned Form A learned a negative scale ($\lambda$ between $-0.26$ and $-0.52$) that reduced attention to all concept-linked tokens almost uniformly (mean bias $-0.176$ on gold and $-0.174$ on distractor tokens); its small F1 gain ($+0.010$, $p=0.11$) was reproduced exactly by the shuffled control. The two features were only weakly predictive of gold concepts (token-level AUC 0.62 for embedding similarity and 0.54 for graph proximity on the test set), which limits what any bias built on them can select.

We therefore replaced them with a stronger relevance signal: a cross-encoder (ms-marco-MiniLM-L-6-v2) scoring each evidence sentence against the question, which identified gold-concept tokens with an AUC of 0.71. Used as an additional Form A feature, it did not help: the query-conditioned Form A again matched its shuffled control (difference $-0.001$, $p=0.65$). Used as the relevance prior of Form B, trained for three epochs, it produced the first concept-specific difference in our experiments. Real concepts reached F1 0.221 and mismatched concepts 0.211, a difference of $+0.010$ (95\% CI $[-0.000,0.024]$, Wilcoxon $p=0.050$). The difference arose from only eight questions, seven improved and one degraded, and it would not remain significant after correction for the number of comparisons reported in this section. To test whether the effect would strengthen with a stronger reader, we repeated this comparison with flan-t5-base (revision \texttt{7bcac57}), placing Form B in encoder layers 8 and 9, the same relative depth as layers 4 and 5 in flan-t5-small, with all other settings unchanged. The larger model answered better without Form B (F1 0.276). Form B raised F1 to 0.281 with real concepts and to 0.281 with mismatched concepts (difference $-0.0003$, 95\% CI $[-0.002,0.001]$, $p=0.66$; 198 of 200 answers identical). The borderline effect observed with flan-t5-small therefore did not replicate, and we conclude that in our experiments neither attention mechanism made measurable use of concept information.

\begin{table}[t]
\centering
\small
\caption{Controls for added trainable capacity (500 training questions, mean over seeds 13, 42, 77). $\Delta$: paired F1 difference to the same model without the mechanism (flan-t5-small: F1 0.188), with 95\% bootstrap CI}
\label{tab:capacity}
\begin{tabular}{lccc}
\hline
\textbf{Configuration} & \textbf{F1} & \textbf{$\Delta$} & \textbf{95\% CI} \\
\hline
Form B, real concepts, 1 epoch & 0.195 & $+0.008$ & $[-0.002, 0.021]$ \\
Form B, mismatched concepts, 1 epoch & 0.195 & $+0.008$ & $[-0.002, 0.021]$ \\
Form B, real concepts, 3 epochs & 0.221 & $+0.033$ & $[0.007, 0.062]$ \\
Form B, mismatched concepts, 3 epochs & 0.213 & $+0.025$ & $[-0.003, 0.056]$ \\
Query-conditioned Form A & 0.197 & $+0.010$ & $[0.000, 0.023]$ \\
Query-conditioned Form A, shuffled & 0.197 & $+0.010$ & $[0.000, 0.023]$ \\
Form B, cross-encoder prior, 3 epochs & 0.221 & $+0.033$ & $[0.008, 0.061]$ \\
\quad same, mismatched concepts & 0.211 & $+0.023$ & $[-0.004, 0.051]$ \\
Query-cond.\ Form A + cross-encoder & 0.197 & $+0.009$ & $[0.000, 0.022]$ \\
\quad same, shuffled features & 0.197 & $+0.010$ & $[0.000, 0.024]$ \\
\hline
\multicolumn{4}{l}{\emph{flan-t5-base (without mechanism: F1 0.276)}} \\
Form B, cross-encoder prior, 3 epochs & 0.281 & $+0.005$ & $[-0.001, 0.013]$ \\
\quad same, mismatched concepts & 0.281 & $+0.005$ & $[-0.001, 0.013]$ \\
\hline
\end{tabular}
\end{table}

\subsection{Computational Efficiency}
\label{subsec:efficiency-results}

The controlled efficiency evaluation compared hybrid graph retrieval with CGA-Form A under identical frozen retrieval, over 200 questions and three repetitions after 20 warm-up queries (Table~\ref{tab:form-a-latency}). Form A added 21.4~ms per query on average (16.0\%), with larger increases at P95 and P99, and no observable change in peak process memory (approximately 2.21~GB). Because Form A did not change any answer, this cost currently buys no output-quality benefit.

\begin{table}[t]
\centering
\small
\caption{Controlled generation latency in milliseconds (CPU, float32)}
\label{tab:form-a-latency}
\begin{tabular}{lcccc}
\hline
\textbf{System} & \textbf{Mean} & \textbf{P50} & \textbf{P95} & \textbf{P99} \\
\hline
Hybrid graph retrieval & 133.9 & 122.6 & 212.8 & 293.6 \\
CGA-Form A & 155.3 & 134.1 & 279.9 & 451.0 \\
\hline
\end{tabular}
\end{table}

\subsection{Temporal Resolution}
\label{subsec:temporal-results}

Table~\ref{tab:temporal} reports results on the 211 LongMemEval questions. All systems received the same retrieved evidence, so evidence recall is identical except for the recency heuristic, which discards earlier sessions. Representing time improved answers substantially: adding dates raised answer accuracy from 0.284 to 0.403 overall, and CLM resolution raised it to 0.379 (CLM versus dense, $+0.095$, 95\% CI $[0.052,0.142]$, $p<10^{-4}$). However, CLM resolution was slightly worse than dated serialisation of the same evidence ($-0.024$, CI $[-0.047,-0.005]$, $p=0.025$). On knowledge-update questions the recency heuristic performed best (0.603), as expected when the latest session always contains the answer, but it discarded the earlier answer-bearing evidence that temporal-reasoning questions require.

The weakness of CLM resolution lies in concept identity. Dense retrieval placed the stale answer first for 19\% of knowledge-update questions, which CLM supersession is designed to correct, but the similarity-based identity rule marked the latest answer-bearing turn as the current version in only 23\% of these questions. When evidence about the same attribute is phrased differently across sessions, it is not recognised as two versions of one concept, and no supersession is recorded.

\begin{table}[t]
\centering
\small
\caption{Temporal resolution on LongMemEval (oracle sessions, flan-t5-large, $k=4$ retrieved user turns). Answer accuracy: normalised gold answer contained in the prediction}
\label{tab:temporal}
\begin{tabular}{lccc}
\hline
\textbf{System} & \textbf{Knowledge update} & \textbf{Temporal reasoning} & \textbf{All} \\
 & ($n=78$) & ($n=133$) & ($n=211$) \\
\hline
Dense (no dates) & 0.410 & 0.211 & 0.284 \\
Dense, dated & 0.551 & 0.316 & 0.403 \\
Recency only & 0.603 & 0.263 & 0.389 \\
CLM resolution & 0.526 & 0.293 & 0.379 \\
\hline
\end{tabular}
\end{table}

\subsection{Epistemic Status}
\label{subsec:epistemic-results}

Table~\ref{tab:epistemic} reports results on the synthetic epistemic graph. All three systems were fine-tuned identically and received the same information: the raw-text system saw every report and the ownership registry, and the graph system saw every claim with its sources. Both reached about 80\% accuracy but presented single-family claims as established facts in most hypothesised cases (84\% and 71\%). They treated two outlets of the same publisher as independent confirmation. With protocol-derived status, accuracy rose to 99.9\% and the unsupported-assertion rate fell from 28.1\% (text) and 23.6\% (graph) to 0.1\%. All systems handled retracted and competing claims well. The difficulty lay specifically in judging source independence, which the CLM confirmation protocol performs explicitly.

\begin{table}[t]
\centering
\small
\caption{Epistemic evaluation on the synthetic graph (500 test instances, 100 per category, disjoint entity names; flan-t5-small fine-tuned on 2{,}000 instances; mean over three seeds). Unsupported: presenting a hypothesised claim as fact, relying on a retracted claim, or preferring an unconfirmed competitor}
\label{tab:epistemic}
\begin{tabular}{lccc}
\hline
 & \textbf{Raw text} & \textbf{Graph, no status} & \textbf{CLM status} \\
\hline
Accuracy & 0.802 & 0.809 & 0.999 \\
\quad validated & 0.853 & 0.757 & 1.000 \\
\quad hypothesised & 0.157 & 0.293 & 1.000 \\
\quad competing & 1.000 & 1.000 & 0.997 \\
\quad invalidated & 1.000 & 1.000 & 1.000 \\
\quad promoted & 1.000 & 0.997 & 1.000 \\
Unsupported-assertion rate & 0.281 & 0.236 & 0.001 \\
\quad on hypothesised claims & 0.843 & 0.707 & 0.000 \\
\hline
\end{tabular}
\end{table}

\subsection{Temporal and Epistemic Results with Larger Models}
\label{subsec:api-results}

Table~\ref{tab:temporal-scale} reports LLM-judged temporal accuracy for four generators. Making time explicit improved accuracy substantially for every model: dated evidence exceeded undated evidence by 12.8 points for flan-t5-large and by 17.1, 25.1, and 20.4 points for the three large models (all $p<10^{-5}$). CLM resolution performed on a par with dated serialisation: the differences were $+1.4$, $+1.9$, and $-1.4$ points for the large models and $-1.4$ for flan-t5-large, none significant. On knowledge-update questions, CLM resolution exceeded dated serialisation for qwen2.5-72b ($+5.1$ points, 95\% CI $[0.013,0.103]$, $p=0.046$) but not for the other models. The benefit of representing time therefore holds, and grows, at scale, whereas the added value of the simplified version resolution over dated evidence remains unestablished.

\begin{table}[t]
\centering
\small
\caption{Temporal resolution on LongMemEval across generators (211 questions, LLM-judged accuracy; $k=4$ retrieved user turns, identical evidence for all systems)}
\label{tab:temporal-scale}
\begin{tabular}{lcccc}
\hline
\textbf{Generator} & \textbf{No dates} & \textbf{Dated} & \textbf{Recency only} & \textbf{CLM} \\
\hline
flan-t5-large (780M) & 0.251 & 0.379 & 0.370 & 0.365 \\
qwen2.5-72b & 0.422 & 0.592 & 0.436 & 0.607 \\
gpt-oss-120b & 0.408 & 0.659 & 0.445 & 0.678 \\
qwen35-122b & 0.408 & 0.611 & 0.408 & 0.597 \\
\hline
\end{tabular}
\end{table}

Table~\ref{tab:epistemic-scale} reports the epistemic evaluation with the three large models. Although the confirmation rule was stated explicitly, all three models presented unconfirmed claims as established facts at high rates when they had to judge source independence themselves: 19--54\% of the relevant cases from raw reports and 66--68\% from claims with their sources. Given claims with their sources but no status, the models also relied on retracted claims in almost every invalidated case. With protocol-derived status, unsupported assertions fell to 0.0\%, 0.3\%, and 5.0\%. gpt-oss-120b tended to answer ``unknown'' for hypothesised claims with status rather than flagging them as unconfirmed, which lowered its accuracy (0.864) without producing unsupported assertions. Model scale therefore does not substitute for explicit epistemic status on this task.

\begin{table}[t]
\centering
\small
\caption{Epistemic evaluation with large models (500 test instances; no fine-tuning; the confirmation rule is stated in the instruction for every input format). Unsupported: presenting a hypothesised claim as fact, relying on a retracted claim, or preferring an unconfirmed competitor}
\label{tab:epistemic-scale}
\begin{tabular}{llccc}
\hline
\textbf{Model} & \textbf{Measure} & \textbf{Raw text} & \textbf{Graph, no status} & \textbf{CLM status} \\
\hline
qwen2.5-72b & Accuracy & 0.798 & 0.566 & 0.986 \\
 & Unsupported & 0.320 & 0.677 & 0.000 \\
gpt-oss-120b & Accuracy & 0.770 & 0.566 & 0.864 \\
 & Unsupported & 0.190 & 0.657 & 0.003 \\
qwen35-122b & Accuracy & 0.586 & 0.590 & 0.920 \\
 & Unsupported & 0.537 & 0.670 & 0.050 \\
\hline
\end{tabular}
\end{table}

\subsection{Summary of Findings}
\label{subsec:results-summary}

The evaluation yields five findings. First, concept-graph retrieval recovers explicit multi-hop structure but does not improve evidence recall. Second, the gold-to-distractor attention preference that appears to show graph-steered attention is a property of the pretrained model; the learned graph bias is about four orders of magnitude smaller than the attention logits and changes no answer. Third, the Form B collapse was caused by a non-identity-preserving normalisation; a repaired Form B preserves the baseline and, trained longer, improves F1 significantly, but most of this gain is reproduced with concepts from other questions, and the concept-specific contribution is not significant. A query-conditioned Form A likewise gains no more than its shuffled control, even with a stronger cross-encoder relevance signal; with that signal as its prior, Form B showed a borderline concept-specific gain with flan-t5-small that did not replicate with flan-t5-base. Fourth, representing time improves temporal question answering substantially for every generator tested, from 780 million to 122 billion parameters, but the simplified CLM version resolution performs on a par with dated serialisation. Fifth, protocol-derived epistemic status nearly eliminates unsupported assertions in a controlled setting where the underlying difficulty is judging source independence, both for a small fine-tuned model and for 72--122-billion-parameter models that are told the rule but must apply it themselves.

%% file: 6-discussion.tex
\section{Discussion}
\label{sec:discussion}

The central hypothesis of this work is that persistent, temporally versioned, and epistemically qualified concepts provide a useful substrate for knowledge-intensive reasoning, and that their structure can be incorporated directly into transformer computation. The results separate this hypothesis into parts that are supported, parts that are not, and parts that the present implementation cannot yet decide.

\subsection{Attention Diagnostics Require a Disabled-Mechanism Control}
\label{subsec:attention-output-gap}

The most consequential finding concerns the apparent evidence for graph-steered attention. With Form A enabled, attention favoured gold over distractor concepts by a factor of 2.76, but the same ratio was obtained with Form A disabled. The pretrained encoder already attends more to token pairs linked to gold concepts, in part because such pairs tend to be close together, and a graph bias of magnitude $2\times10^{-3}$ added to attention logits with a standard deviation above four cannot change this pattern measurably. Comparing gold with distractor pairs within the Form A condition alone, without the counterfactual measurement, would have attributed this preference to Form A.

This is a general hazard for knowledge-enhanced attention mechanisms. Whenever the knowledge source is correlated with what a pretrained model already attends to, as gold evidence usually is, an attention map that favours the knowledge-linked tokens is expected with or without the mechanism. Attention diagnostics should therefore always be paired with the same measurement under the disabled mechanism, and ideally with output-level interventions, in line with earlier cautions about interpreting attention weights \cite{jain2019attention,serrano2019attention,wiegreffe2019attention}.

The scaling experiment shows that the problem is not only that the learned bias is small. Increasing it by two orders of magnitude changed a handful of answers without improving F1, and increasing it further degraded generation. Moreover, the gold-to-distractor ratio did not rise at any scale. Because the bias is added to every concept-linked token pair in proportion to edge strength and path proximity, distractor pairs, which are also connected in a dense co-occurrence graph, receive a comparable share. An additive bias of this form is therefore both too weak at the learned scale and insufficiently selective at larger scales. We therefore tested a bias defined relative to the query, which boosts tokens of concepts similar or close in the graph to the question, including a cross-encoder relevance score. It learned instead to reduce attention to all concept tokens uniformly, and its small gain was reproduced with shuffled features in every configuration. The available relevance signals identify gold concepts only weakly (AUC about 0.6), so the bottleneck lies less in how structure enters attention than in determining which concepts matter for a question.

\subsection{What the Form B Failure Was, and What It Was Not}
\label{subsec:form-b-discussion}

The collapse of Form B was an implementation failure with a precise cause: a layer normalisation applied to the residual stream of a pre-normalisation transformer. It destroyed the pretrained computation before any concept information was used, and a gate forced to zero did not prevent it. The dual-attention composition in Section~\ref{subsec:cga} is formulated to exclude this failure. Once the layer was made identity-preserving, Form B was harmless at initialisation and, after training, attended 3.5 times more to each gold concept than to each distractor. Trained for three epochs, it raised F1 significantly, from 0.188 to 0.221. The capacity control shows, however, that concepts taken from other questions raised F1 to 0.213, so most of the gain came from added trainable capacity rather than from concept information, and the concept-specific contribution of 0.008 was not significant. With a cross-encoder relevance prior the concept-specific contribution rose to 0.010, at the margin of significance ($p=0.05$) and based on eight questions. This effect did not replicate with flan-t5-base, where real and mismatched concepts produced the same answers for 198 of 200 questions. We therefore find no reliable evidence that either attention mechanism uses concept information at the scales we could test.

The diagnosis also corrects the interpretation that Form B attended mainly to distractor concepts. Total attention mass on distractors exceeded mass on gold concepts only because distractors were more numerous. Per concept, Form B favoured gold concepts, but this preference followed from the concept features and relevance prior rather than from the question, since it was unchanged when concepts were swapped between questions. The remaining limitation is how much this selectivity can add when the concept representations are short labels whose information is largely present in the retrieved passages, and when the generator is small. Form B is designed to supply information absent from the text; evaluating it where concept nodes carry such information, for example aggregated evidence from passages outside the context window, is a necessary next step.

\subsection{Temporal Information Helps; Version Resolution Needs Concept Identity}
\label{subsec:temporal-discussion}

On LongMemEval, making time explicit improved answers by roughly ten points for flan-t5-large and by 17 to 25 points for three models of 72 to 122 billion parameters, which supports the premise that temporal validity should be represented rather than left implicit, and shows that stronger readers exploit it more rather than less. CLM resolution matched this gain but did not exceed simply presenting the same evidence with dates, apart from a small advantage on knowledge-update questions for one model. The analysis locates the shortfall in concept identity: supersession can only be recorded when two pieces of evidence are recognised as versions of the same concept, and an embedding-similarity rule over conversational turns rarely achieves this. The lifecycle model assumes reliable concept identity; the experiment shows that this assumption carries most of the weight, and that a version-resolution layer adds value only if identity is established at the level of the attribute being updated rather than the surface similarity of the evidence.

\subsection{Epistemic Status Is Useful Where Independence Must Be Judged}
\label{subsec:epistemic-discussion}

On the synthetic graph, models that saw the same reports and ownership registry treated two outlets of one publisher as independent confirmation in most cases, and therefore presented single-source claims as facts. Protocol-derived status removed nearly all such unsupported assertions. The result should be interpreted with care. The confirmation rule computes exactly the property on which the test categories are defined, so the experiment demonstrates that making this computation explicit and passing its result to the model is effective, and that models do not reliably perform it themselves: neither a small fine-tuned model nor 72--122-billion-parameter models that were told the rule explicitly, which still presented single-source claims as established in 19--68\% of the relevant cases. It does not show that the protocol's notion of independence is correct in real corpora, where common provenance is often hidden and ownership registries are incomplete.

\subsection{Retrieval-Level Value of Concept Graphs}
\label{subsec:retrieval-value}

Concept-graph retrieval exposed explicit paths between distributed evidence, which passage retrieval cannot provide, and hybrid retrieval improved answer F1 over dense retrieval. It did not, however, improve recall of supporting passages, and serialising the retrieved paths did not help the generator. In the present setting, the graph's clearest value is inspectability and provenance rather than accuracy.

\subsection{Limitations}
\label{subsec:limitations}

The multi-hop experiments use generators of 77 and 250 million parameters with a 384-token budget, and all systems score low in absolute terms; effects that require substantially larger readers may not appear at these scales. The multi-hop test set contains 200 questions, which limits the power to detect small differences such as the Form B improvement. The temporal evaluation uses the oracle sessions of one benchmark, and its LLM judge (gpt-oss-120b) is also one of the evaluated generators, so its scores for that model could be biased; the containment metric gives the same qualitative picture. The CLM resolution rule is a simplified instance of the lifecycle model. The large models were evaluated through an inference endpoint, so Forms A and B, which require modifying and training layers inside the model, could not be tested at that scale. The epistemic evaluation is synthetic, and its categories are defined by the same rule that the CLM applies. The adjudicator, the full lifecycle dynamics of emergence, drift, and splitting, and scaling to very large graphs remain unevaluated. 

\subsection{Future Work}
\label{subsec:future-work}

Three directions follow directly from the results. First, structural attention should be conditioned on query relevance and evaluated with disabled-mechanism controls and larger generators. Second, Form B should be evaluated where concept nodes carry information absent from the textual context, with more training data. Third, concept identity for version resolution should be established at the attribute level, for example by extracting subject--attribute--value records, before supersession is applied. Broader comparisons with graph-based retrieval systems \cite{edge2024graphrag,gutierrez2024hipporag,mavromatis2025gnnrag} and with structure-aware retrievers on hierarchical benchmarks such as SearchTome \cite{kumar2026stair}, which would require adapting CGA to retrieval over table-of-contents hierarchies, would position attention-level structural injection against retrieval-level and prompt-level alternatives.

%% file: 7-conclusion.tex
\section{Conclusion}\label{sec:conclusion}

This paper introduced the Concept Lifecycle Model (CLM), which formalises concepts as persistent, graph-grounded, temporally versioned entities with explicit provenance and epistemic status, and Concept-Grounded Attention (CGA), which incorporates concept-graph information into transformer computation through graph-biased self-attention (Form A) and gated cross-attention over concept nodes (Form B). We evaluated the framework in three controlled settings, pairing every internal diagnostic with the same measurement under a disabled mechanism.

The results are mixed and specific. Concept-graph retrieval recovered explicit multi-hop structure but did not improve evidence recall. The gold-to-distractor attention preference that appears to show graph-steered attention is a property of the pretrained model: it is identical with Form A disabled, and the learned bias is about four orders of magnitude smaller than the attention logits. The failure of Form B was caused by a normalisation that did not preserve the pretrained computation; a repaired Form B improved answers, but mostly through added trainable capacity, as concepts from other questions produced most of the same gain; a borderline concept-specific gain with a strong relevance prior did not replicate with a larger model, and a query-conditioned Form A gained no more than a shuffled control. Representing time improved temporal question answering for every generator tested, by 13 to 25 points and most strongly for the largest models, but simplified CLM version resolution did not outperform dated serialisation, because concept identity was not established reliably. Protocol-derived epistemic status nearly eliminated unsupported assertions in a controlled setting where source independence must be judged, and models of 72 to 122 billion parameters that were told the rule could not replace it, still asserting unconfirmed claims in 19 to 68 percent of cases.

Taken together, the findings support the representational premises of the CLM, that temporal validity and epistemic status should be explicit, more than they support the proposed mechanisms for injecting graph structure into attention. They also yield three methodological lessons for knowledge-enhanced language models: attention diagnostics are uninformative without a disabled-mechanism control when the knowledge source correlates with what the model already attends to; modules inserted into pretrained networks must preserve the pretrained computation at initialisation; and gains from trainable knowledge modules must be compared with the same module given mismatched knowledge.

Future work will condition structural attention on query relevance and evaluate it with larger generators, evaluate concept cross-attention where concept nodes carry information absent from the text, and establish attribute-level concept identity before applying version resolution. The CLM and CGA remain a concrete and testable framework for separating persistent knowledge representation from language generation; the present study identifies which of its components currently deliver on that separation and which do not.

%% file: 8-appendix.tex
\section{Formal Specification of Unevaluated Components}
\label{app:unevaluated}

This appendix gives the complete formulation of the components of the Concept Lifecycle Model and CGA pipeline that are summarised in Section~\ref{sec:methodology} and were not evaluated, or were evaluated only in simplified form, in the present study. They are included to make the framework fully specified and to support future empirical evaluation.

\subsection{Temporal and Epistemic Annotation}
\label{app:temporal-epistemic-annotation}

Each node and relationship is associated with a temporal record
\begin{equation}
    T_{ij}
    =
    \left(
        t_{\mathrm{valid\_from}},
        t_{\mathrm{valid\_to}},
        t_{\mathrm{recorded}},
        t_{\mathrm{updated}}
    \right).
\end{equation}

The valid-time interval describes when the concept or relationship is considered true in the represented domain. The transaction-time fields describe when the system first recorded and subsequently modified the item. Open-ended current relationships use
\begin{equation}
    t_{\mathrm{valid\_to}}=+\infty.
\end{equation}

When contradicting evidence arrives, the current validity interval is closed and an invalidation event is created. The previous state remains queryable for temporal anchors within its historical interval.

Epistemic annotation is based on the confirmation path associated with each edge. The provenance set is partitioned into source families
\begin{equation}
    P_{ij}
    =
    \mathcal{F}_{ij}^{(1)}
    \cup
    \mathcal{F}_{ij}^{(2)}
    \cup
    \cdots
    \cup
    \mathcal{F}_{ij}^{(m)},
\end{equation}
where evidence items belonging to the same upstream origin, republication chain, or near-duplicate family are grouped together.

A relationship may be validated when the confirmation rule
\begin{equation}
    \operatorname{Confirm}(e_{ij})
    =
    \mathbb{I}
    \left[
        m \geq n_{\mathrm{indep}}
        \;\lor\;
        h_{ij}=1
        \;\lor\;
        s_{ij}=1
    \right]
\end{equation}
is satisfied, where $n_{\mathrm{indep}}$ is the required number of independent source families, $h_{ij}$ indicates human confirmation, and $s_{ij}$ indicates direct confirmation from an accepted structured source.

This rule is configurable by domain. High-stakes settings may require stricter source-independence conditions or mandatory human review, while lower-risk applications may use more permissive confirmation thresholds.

\subsection{Temporally Scoped Retrieval and Adjudication}
\label{app:routing}

\subsubsection*{Path C: Temporally Scoped Retrieval}

Path C is activated when the query contains an explicit temporal anchor or when the adjudicator determines that current-state retrieval is insufficient. It resolves each candidate concept and relationship to the state valid at $t_q$:
\begin{equation}
    C_i^{t_q}
    =
    \operatorname{Resolve}(C_i,t_q).
\end{equation}

The temporal retrieval result is
\begin{equation}
    \mathcal{R}_C
    =
    \left\{
        C_i^{t_q},
        e_{ij}^{t_q}
        \mid
        \operatorname{relevant}
        \left(
            C_i^{t_q},
            q
        \right)=1
    \right\}.
\end{equation}

Path C distinguishes valid time from transaction time. Evidence recorded after $t_q$ may still be used when it retrospectively describes a state valid at $t_q$, provided that the transaction-time policy permits retrospective knowledge. Conversely, evidence valid only after $t_q$ is excluded even when it is highly semantically similar to the query.

The representative temporal embedding is
\begin{equation}
    \overline{\mathbf{r}}_C
    =
    \sum_{C_i^{t_q}\in\mathcal{R}_C}
    s_i^{C}
    \mathbf{z}_i^{t_q}.
\end{equation}

\subsubsection*{Adjudicator Scoring}

For each retrieval path $p\in\{A,B,C\}$, the adjudicator receives a feature vector
\begin{equation}
    \mathbf{u}_p
    =
    \left[
        \cos
        \left(
            \mathbf{q},
            \overline{\mathbf{r}}_p
        \right),
        \operatorname{coverage}_p,
        \operatorname{epistemic}_p,
        \operatorname{temporal}_p,
        \operatorname{cost}_p
    \right].
\end{equation}

The features represent semantic relevance, concept coverage, average epistemic quality, temporal compatibility, and estimated computational cost.

A confidence score is produced through
\begin{equation}
    a_p
    =
    \operatorname{softmax}
    \left(
        f_{\mathrm{adj}}
        \left(
            \mathbf{u}_A,
            \mathbf{u}_B,
            \mathbf{u}_C
        \right)
    \right)_p.
\end{equation}

The adjudicator can operate in either selection or fusion mode. Selection is used when one retrieval path has substantially greater confidence:
\begin{equation}
    p^{*}
    =
    \underset{p}{\arg\max}\;a_p.
\end{equation}

In this case,
\begin{equation}
    \mathcal{G}_{q,t,P}
    =
    \mathcal{R}_{p^{*}}.
\end{equation}

Fusion is used when multiple paths contain complementary evidence:
\begin{equation}
    \mathcal{G}_{q,t,P}
    =
    \operatorname{Prune}_{K_{\max}}
    \left(
        \bigcup_{p\in\mathcal{P}^{*}}
        \mathcal{R}_p
    \right),
\end{equation}
where
\begin{equation}
    \mathcal{P}^{*}
    =
    \left\{
        p:
        a_p\geq\tau_{\mathrm{fuse}}
    \right\}.
\end{equation}

Duplicate concepts are merged through their persistent identifiers. Conflicting relationships are not averaged blindly; they are preserved with their temporal intervals, provenance, and epistemic statuses. The resulting subgraph is ranked and pruned to satisfy $K_{\max}$.

\subsubsection*{Sequential Early Stopping}

To reduce unnecessary retrieval cost, the paths are executed sequentially.

Path A is executed first. If
\begin{equation}
    a_A
    \geq
    \tau_A
\end{equation}
and the query does not contain an unresolved temporal or multi-hop requirement, retrieval terminates.

Otherwise, Path B is executed. If
\begin{equation}
    \max(a_A,a_B)
    \geq
    \tau_B
\end{equation}
and no temporal anchor is present, the adjudicator selects or fuses Paths A and B.

Path C is executed when:

\begin{itemize}
    \item the query contains an explicit temporal expression;
    \item the query requests change over time;
    \item conflicting concept versions are detected;
    \item confidence after Paths A and B remains below the stopping threshold.
\end{itemize}

The procedure can be summarized as
\begin{equation}
    \mathcal{R}
    =
    \begin{cases}
        \mathcal{R}_A,
        &
        a_A\geq\tau_A,
        \\[4pt]
        \operatorname{Adjudicate}
        (\mathcal{R}_A,\mathcal{R}_B),
        &
        a_A<\tau_A
        \land
        \neg\operatorname{Temporal}(q),
        \\[4pt]
        \operatorname{Adjudicate}
        (\mathcal{R}_A,\mathcal{R}_B,\mathcal{R}_C),
        &
        \text{otherwise}.
    \end{cases}
\end{equation}

\subsubsection*{Cold-Start Routing}

Before sufficient labelled routing data are available, the adjudicator uses a deterministic warm-start policy. Queries containing explicit dates or expressions such as ``as of'', ``before'', ``after'', or ``between'' activate Path C. Queries containing multiple entities, relational expressions, comparisons, or causal terms activate Path B. Remaining queries begin with Path A.

After accumulating routing examples, the learned adjudicator gradually replaces the rule-based policy. Let $N$ denote the number of observed training queries. A mixture coefficient
\begin{equation}
    \lambda_{\mathrm{route}}
    =
    \min
    \left(
        1,
        \frac{N}{N_{\mathrm{bootstrap}}}
    \right)
\end{equation}
controls the transition:
\begin{equation}
    a_p^{\mathrm{final}}
    =
    \lambda_{\mathrm{route}}
    a_p^{\mathrm{learned}}
    +
    \left(
        1-\lambda_{\mathrm{route}}
    \right)
    a_p^{\mathrm{rule}}.
\end{equation}

\subsection{Temporal, Epistemic, and Routing Objectives}
\label{app:objectives}

\subsubsection*{Epistemic Calibration Objective}

The epistemic objective encourages the model to distinguish evidence quality and avoid presenting hypothesised paths as validated facts. Let
\begin{equation}
    z_r
    \in
    \{
        \textit{validated},
        \textit{hypothesised},
        \textit{invalidated}
    \}
\end{equation}
denote the target epistemic class associated with generated claim $r$.

The calibration loss is
\begin{equation}
    \mathcal{L}_{\mathrm{epi}}
    =
    -
    \sum_r
    \log
    p
    \left(
        z_r
        \mid
        q,
        \mathcal{G}_{q,t,P}
    \right).
\end{equation}

For invalidated or temporally incompatible evidence, an additional margin loss can penalize attention allocation:
\begin{equation}
    \mathcal{L}_{\mathrm{invalid}}
    =
    \sum_{j\in\mathcal{C}_{\mathrm{invalid}}}
    \max
    \left(
        0,
        a_j-\varepsilon_{\mathrm{inv}}
    \right),
\end{equation}
where $a_j$ is the total attention assigned to invalid evidence.

\subsubsection*{Temporal Objective}

The temporal loss encourages selection of the concept state valid at the requested time. Let
\begin{equation}
    v_i^{*}
\end{equation}
be the correct version of concept $C_i$ for temporal anchor $t_q$. The version-selection loss is
\begin{equation}
    \mathcal{L}_{\mathrm{temp}}
    =
    -
    \sum_i
    \log
    p
    \left(
        v_i^{*}
        \mid
        C_i,
        t_q
    \right).
\end{equation}

Negative temporal examples include future states, expired relationships, pre-birth concepts, and facts that were recorded before the query but were not yet valid.

\subsubsection*{Adjudicator Objective}

Let
\begin{equation}
    p^{*}
    \in
    \{A,B,C,\mathrm{fusion}\}
\end{equation}
denote the optimal retrieval strategy for a training query. The routing loss is
\begin{equation}
    \mathcal{L}_{\mathrm{route}}
    =
    -
    \log
    p
    \left(
        p^{*}
        \mid
        \mathbf{u}_A,
        \mathbf{u}_B,
        \mathbf{u}_C
    \right).
\end{equation}

A computational-cost regularizer may be added:
\begin{equation}
    \mathcal{L}_{\mathrm{cost}}
    =
    \sum_{p}
    a_p c_p,
\end{equation}
where $c_p$ is the normalized cost of retrieval path $p$. This encourages the adjudicator to select the least expensive sufficient path rather than always executing the full retrieval pipeline.